\documentclass[sigplan,screen]{acmart}

\AtBeginDocument{%
  \providecommand\BibTeX{{%
    \normalfont B\kern-0.5em{\scshape i\kern-0.25em b}\kern-0.8em\TeX}}}

\renewcommand\footnotetextcopyrightpermission[1]{} 
\setcopyright{acmcopyright}
\copyrightyear{2022}
\acmYear{2022}
\acmDOI{XXXXXXX.XXXXXXX}

\usepackage{multicol}
\usepackage{multirow}
\begin{document}

\title{Adaptive Pattern Extraction Multi-Task Learning for Multi-Step Conversion Estimations}

\author{Xuewen Tao}
\authornote{These authors contributed equally to this research.}
\email{xuewen.txw@mybank.cn}
\affiliation{%
  \institution{MYbank, Ant Group}
  \city{Beijing}
  \country{China}
}

\author{Mingming Ha}
\authornotemark[1]
\email{hamingming_0705@foxmail.com}
\affiliation{%
  \institution{School of Automation and Electrical Engineering, University of Science and Technology Beijing; MYbank, Ant Group}
  \city{Beijing}
  \country{China}
}

\author{Qiongxu Ma}
\email{qiongxu.mqx@mybank.cn}
\affiliation{%
  \institution{MYbank, Ant Group}
  \city{Shanghai}
  \country{China}
}

\author{Hongwei Cheng}
\email{chw286885@mybank.cn}
\affiliation{%
  \institution{MYbank, Ant Group}
  \city{Shanghai}
  \country{China}
}

\author{Wenfang Lin}
\email{moxi.lwf@mybank.cn}
\affiliation{%
  \institution{MYbank, Ant Group}
  \city{Hangzhou}
  \state{Zhejiang}
  \country{China}
}

\author{Xiaobo Guo}
\authornote{Xiaobo Guo is the corresponding author.}
\email{xb_guo@bjtu.edu.cn}
\affiliation{%
  \institution{Institute of Information Science, Beijing Jiaotong University, Mybank, Ant Group,}
  \city{Beijing}
  \country{China}
}

\begin{abstract}
Multi-task learning (MTL) has been successfully used in many real-world applications, which aims to simultaneously solve multiple tasks with a single model. The general idea of multi-task learning is designing kinds of global parameter sharing mechanism and task-specific feature extractor to improve the performance of all tasks. However, challenge still remains in balancing the trade-off of various tasks since model performance is sensitive to the relationships between them. Less correlated or even conflict tasks will deteriorate the performance by introducing unhelpful or negative information. Therefore, it is important to efficiently exploit and learn fine-grained feature representation corresponding to each task. In this paper, we propose a 
Adaptive Pattern Extraction Multi-task (APEM) framework, which is adaptive and flexible for large-scale industrial application. APEM is able to fully utilize the feature information by learning the interactions between the input feature fields and extracted corresponding tasks-pecific information. We first introduce a DeepAuto Group Transformer module to automatically
and efficiently enhance the feature expressivity with a modified set attention mechanism and a Squeeze-and-Excitation operation. Second, explicit Pattern Selector is introduced to further enable selectively feature representation learning by adaptive task-indicator vectors.
Empirical evaluations show that APEM outperforms the state-of-the-art MTL methods on public
and real-world financial services datasets. More importantly, we explore the online performance
of APEM in a real industrial-level recommendation scenario.
\end{abstract}






\maketitle

\section{Introduction}\label{introduction}
\begin{figure}[htbp]
\centering
	\includegraphics[width=0.8\linewidth]{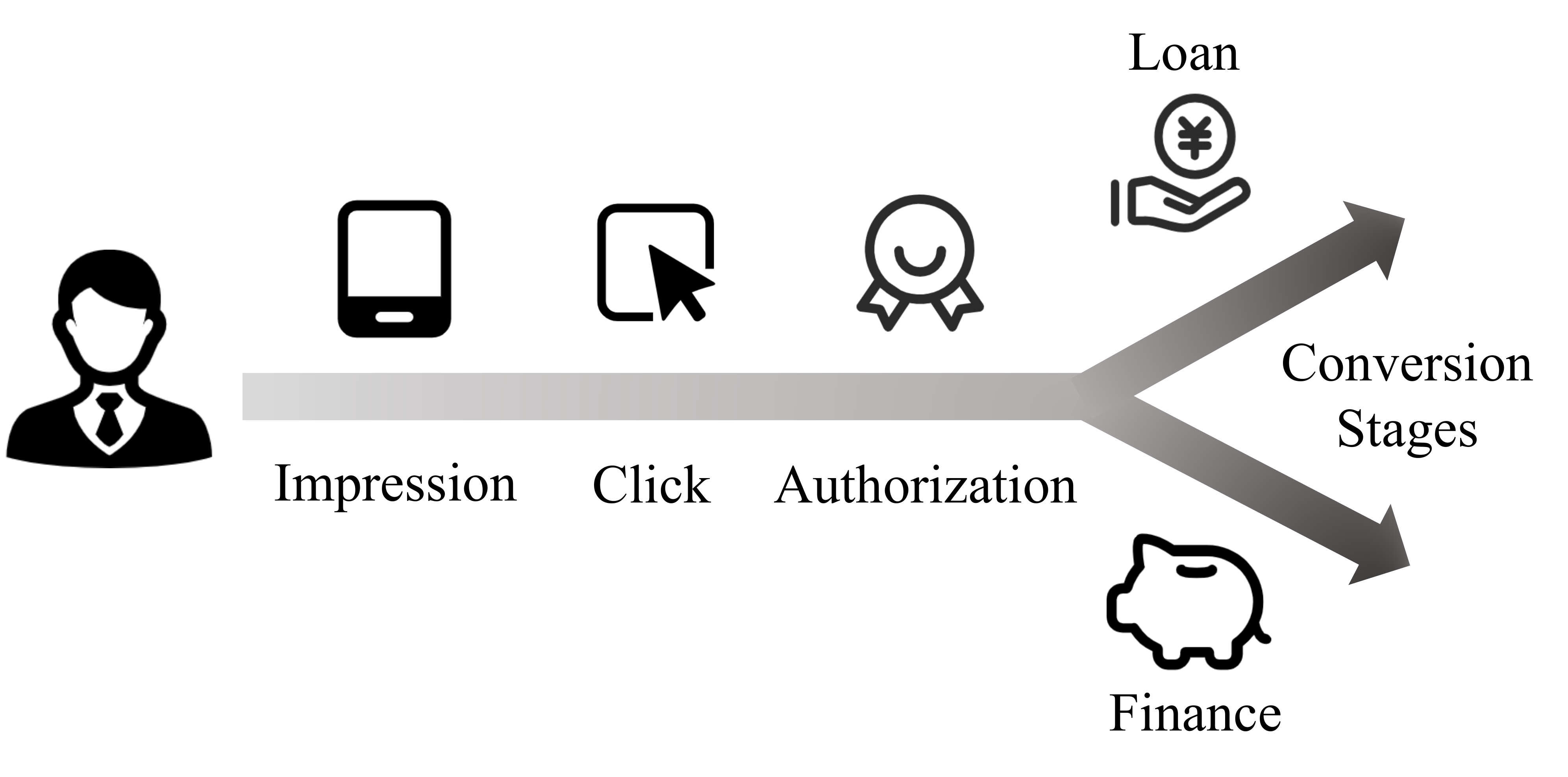}
	\caption{An illustration of a multi-step conversion in Finance Service.}
	\label{intro}
\end{figure}

Multi-task learning (MTL) has enjoyed rather remarkable successes for various real-world scenarios, such as the online recommendation \cite{guo2017deepfm,tang2020progressive}, display advertising \cite{fei2021gemnn}, customer acquisition management, financial service \cite{AITM2021} and so forth. MTL techniques simultaneously learn multiple tasks by implicitly passing the message among related tasks \cite{vandenhende2020revisiting,Chen2021IS}, compared with single-task learning, which can improve the overall performance of these tasks \cite{shen2021variational,crawshaw2020multi}. In online advertising,
recommendation, and customer acquisition etc, post-view click-through rate (CTR), post-click conversion rate (CVR) and post-view click-through \& conversion rate (CTCVR) estimations are a series of classical tasks \cite{ESMM2018,Entire_SIGIR_2020} with sequential dependence of the customer acquisition process. In this case, sequential pattern of user behaviours means the later action only occur after the former action. More generally, this sequential pattern can be extended to multi-step conversion. As shown in Fig. \ref{intro}, a multi-step conversion example in fiance service strictly follow this sequential dependence pattern. A customer will convert through stages of \textit{Impression} $\to$ \textit{Click} $\to$ \textit{Authorize} $\to$ \textit{Conversion}. Conversion behaviors such like applying loans, make deposit or purchasing investment products are only permitted after an authorization. Aimed at various kinds of concrete industrial applications, extensive studies \cite{ma2018modeling,ESMM2018,Micro_Macro2021,Causal_ESCM2_2022,Causal_Estimating_2021,wang2022enhancing} focus on the CTR and CVR estimations. However, there are rare works to provide a formalized definition of the sequential dependence MTL (SDMTL) problem, which is of particular significance in multi-step conversion estimations applicable to more diverse scenarios. In addition, the connection and difference between the general MTL and SDMTL is also unclear. 

Series works from ESMM \cite{ESMM2018} to $\text{ESCM}^2$ \cite{Entire_SIGIR_2020} pay more attention on the unbiased CVR estimation problem in a view of causality to correct sample selection bias. The dependent relation between the tasks like CTR and CVR is implicitly implied via the distribution of sample space. Recently, \cite{AITM2021} captures the task dependency through the information transfer between different conversion steps and combines a calibrator to further constrain the dependent relationship. However, the dependency between each steps is still not deeply defined and discussed in most MTL works from a theoretical format. 

Besides, as mentioned above, MTL methods improve prediction result through an information passing mechanism between tasks, which suggests improper feature sharing will result in even poorer or imbalanced performance in different tasks known as a negative transfer phenomenon. Therefore, general approach in MTL mainly focuses on designing kinds of information extraction modules (experts) to learn common and task-specific representations. Such as Cross-Stitch Network~\cite{misra2016cross} and Sluice Network~\cite{ruder2019latent} employ a linear combination to leverage representations of different tasks but also require much more training parameters. SOTA method of Multi-gate Mixture-of-Experts (MMoE) approach ~\cite{ma2018modeling} adopts an ensemble of experts submodules and gating network to model task relationships while consuming less computation. Progressive Layered Extraction (PLE) ~\cite{tang2020progressive}, separates task-common and task-specific parameters explicitly which could further avoid parameter conflicts caused by complex task correlation. These approaches assign individual parameters to each task to better exploit task information and improve model generalization. Nonetheless, feature expressivity with respect to each task is still limited since task-irrelevant information passing from the shared structure and more fine-grained representation learning is necessary.


In this paper, we first provide a formal definition of MTL on sequential dependence problem, and propose an optimizing object paradigm for recover the dependent relationship based on theoretical proof. And we also present a novel MTL framework called Adaptive Pattern Extraction Multi-task (APEM) framework to selectively and dynamically enhance the representation learning for respective tasks along with the dependency-based object.
APEM consists of two main modules: Adaptive Sample-wise Representation Generator (ASRG) and explicit Pattern Selector (PS). ASRG employs a dynamic selection mechanism 
 to learn the hierarchical feature interaction from a sample-wise view to further separates the task-irrelevant information. The implement of explicit PS enables fine-grained feature learning by introducing task-specific indicator vectors. 
In a summary,  main contributions of this paper are presented as follows:
	\begin{itemize}
	   \item 
        The SDMTL problem is first formally formulated, and its connections and differences with the general MTL problem are illustrated. Moreover, the distribution dependence relationship between the adjacent task spaces is revealed from a theoretical perspective.
		\item
		We present a multi-task learning framework named APEM for selectively fine-grained feature representation learning from a sample-wise view. ASRG and PS modules within APEM adatively reconstruct the implicit shared representations and extract explicit task-specific information in an more efficient way. 
  
		
		\item
	Extensive experiments on public and real-world industrial dataset are conducted to evaluate the effectiveness of APEM. Experiment results demonstrate that our proposed approach outperforms the state-of-the-art MTL methods. Furthermore, we explore the boundary of APEM in real-world industrial applications to prove its efficiency for large-scale online recommendations.
	\end{itemize}

\section{Preliminaries}
In this section, the SDMTL problem and the connection between SDMTL
and general MTL are elaborated. Then, from the expected loss's point of view, the distribution relationship between the adjacent task domains is revealed.

\subsection{Problem Formulation}\label{ProblemFormulation}
Consider a SDMTL problem over an input space $\mathbb{X}$ and a set of task $\{\mathcal{T}_i\}_{i=1}^N$, where $N$ is the number of tasks and the corresponding task spaces are denoted as $\{\mathbb{T}_1,\dotsc,\mathbb{T}_N\}$. A large dataset of data points $\{x_j,o_j^1,\dotsc,o_j^N\}_{j=1}^{M}$ are given, where $M$ is the number of data points and $o_i^j\in\{0,1\}$ corresponding to a binary classification problem or $o_i^j\in\mathbb{R}$ for a regression problem is the label of the $i$-th task for the $j$-th data point. Differing from the general MTL problem, for the SDMTL problem, there exists the sequential dependence relationship between tasks in the sense that the current task $\mathcal{T}_i$ depends on the previous task $\mathcal{T}_{i-1}$, i.e., $\mathcal{T}_{i-1}\to\mathcal{T}_{i}$. Let $X$ and $T_i$ be the random variables over the input space $\mathbb{X}$ and task output space $\mathbb{T}_i$, respectively. In this paper, for convenience of analysis, each task is set as a binary classification task. As mentioned in literature \cite{o2021analysis} and shown in Fig. \ref{intro}, for sequential dependence, one of core properties is that if the event $T_{i-1}$ is not triggered, then the event $T_i$ must not occur, i.e., $P(T_i=1,T_{i-1}=0\vert{X})=0$, where $P(\cdot\vert\cdot)$ denotes the conditional probability. Therefore, according to this property, the random variables $T_i$ satisfies 
\begin{align}
\label{Eq2.1_01}
P(T_i=1\vert{X})=&\sum_{t_{i-1},\dotsc,{t_1}\in\{0,1\}}P(T_i=1,\dotsc,{T_1=t_1}\vert{X})\nonumber\\
=&\;P(T_i=1, T_{i-1}=1,\dotsc,{T_1=1}\vert{X})\nonumber\\
=&\;P(T_i=1, T_{i-1}=1\vert{X}),\nonumber\\
P(T_i=0\vert{X})=&\sum_{t_{i-1},\dotsc,{t_1}\in\{0,1\}}P(T_i=0,\dotsc,{T_1=t_1}\vert{X})\nonumber\\
=&\;P(T_i=0,T_{i-1}=0\vert{X})+P(T_i=0,T_{i-1}=1\vert{X}),
\end{align}
which implies that the positive samples of $T_i$ are derived from the positive samples of the task $T_i$ while the negative samples consist of the negative samples of tasks $T_{i}$ and $T_{i-1}$ due to the sequential dependence.

In addition, the sequential dependence relationship is also embodied in the constraints with respect to the conversion probabilities of the adjacent tasks. In \cite{AITM2021}, the sequential dependence relationship is formalized as
\begin{align}
\label{Eq2.1_02}
P(T_1=1\vert{X})\ge&{P(T_2=1,T_1=1\vert{X})}\nonumber\\
&\cdots\nonumber\\
\ge&P(T_{i-1}=1,\dotsc,T_1=1\vert{X})\nonumber\\
\ge&{P}({T}_i=1,T_{i-1}=1,\dotsc,T_1=1\vert{X}).
\end{align}
Then, a behavioral expectation calibrator is introduced into AITM \cite{AITM2021} to guarantee the sequential dependence relationship (\ref{Eq2.1_02}). When the outputs of the model violate this condition, the designed loss will output a positive penalty term. However, the condition (\ref{Eq2.1_02}) cannot completely reflect the dependence relationship between tasks. Reconsidering the dependence relationship between $P(T_{i-1}=1,\dotsc,T_1=1\vert{X})$ and $P(T_i=1,\dotsc,T_1=1\vert{X})$, it leads to 
\begin{align}
\label{Eq2.1_03}
P(T_{i-1}=1\vert{X})-&P(T_i=1\vert{X})\nonumber\\
&=P(T_{i-1}=1\vert{X})-P(T_i=1,T_{i-1}=1\vert{X})\nonumber\\
&=P(T_{i-1}=1\vert{X})\big[1-P(T_i=1\vert{T_{i-1}=1,X})\big]\nonumber\\
&=P(T_{i-1}=1\vert{X})P({T}_i=0\vert{T_{i-1}=1,X})\nonumber\\
&=P({T}_i=0,T_{i-1}=1\vert{X}).
\end{align}
Therefore, the dependence relationship between the adjacent tasks needs to satisfy the equality constraints (\ref{Eq2.1_03}), which also contains the dependence information $P(T_i=1,T_{i-1}=0\vert{X})=0$.

Define a parametric hypothesis class per task as $f_i(x;\theta^{s},\theta^{i})\colon\\
\mathbb{X}\to\mathbb{T}_i$, where $\theta^{s}$ and $\theta^{i}$ are shared parameters and task-specific parameters of the task $i$. Also, the task-specific loss function is defined as $L_i(\cdot,\cdot)\colon\mathbb{T}_i\times\mathbb{T}_i\to\mathbb{R}^+$.
Similar to the general MTL problem, the objective of SDMTL is to minimize the following expected loss:
\begin{align}
\label{Eq2.1_04}
\min_{\theta_s,\theta_1,\dotsc,\theta_N}\sum_{i=1}^N&E_{X,T_1,\dotsc,T_N\thicksim\mathcal{O}}[w_iL_i(f_i(X;\theta^{s},\theta^{i}),T_i)]\nonumber\\
\text{s.t.}\;f_i(X;\theta^{s},\theta^{i})&-f_{i-1}(X;\theta^{s},\theta^{i})=P({T}_i=0,T_{i-1}=1\vert{X}),\nonumber\\
&\quad\quad\quad\quad\quad\quad\quad\quad\quad\quad\quad{i}=2,\dotsc,N,
\end{align}
where $\mathcal{O}$ is the distribution with domain $\mathbb{X}\times\mathbb{T}_1\times\cdots\times\mathbb{T}_N$, and $w_i$ is the static or dynamically computed weight per task. Therefore, the SDMTL problem can be considered as a general MTL with the constraints $f_{i-1}(x_j;\theta^s,\theta^{i-1})-f_i(x_j;\theta^s,\theta^i)=P({T}_i=0,T_{i-1}=1\vert{X})$, which implies that the difference of $f_{i-1}(x_j;\theta^s,\theta^{i-1})$ and $f_i(x_j;\theta^s,\theta^i)$ is the probability of the event $T_i$ not occurring when the event $T_{i-1}$ is triggered. Considering the binary classification tasks and the labels $o^{i}_j$ and $o^{i-1}_j$ of the adjacent tasks, we can obtain the label corresponding to the probability $P({T}_i=0,T_{i-1}=1\vert{X})$ as 
\begin{table}[htbp]  
\caption{\label{SequentialDependence}Labels corresponding to tasks $\mathcal{T}_{i-1}$, $\mathcal{T}_{i}$, and the dependence relationship.}
\begin{tabular}{cccc}    
\toprule    
$o^{i}_j$ & $o^{i-1}_j$ & $o^{i}_j-o^{i-1}_j$ & $P({T}_i=0,T_{i-1}=1\vert{X})$\\    
\midrule   
0 & 0 & 0 & 0\\   
1 & 0 & 1 & 1\\   
1 & 1 & 0 & 0\\    
\bottomrule   
\end{tabular}  
\end{table}

Therefore, according to Table \ref{SequentialDependence}, the label of the sequential
dependence between the adjacent tasks is equivalent to the difference between $o^{i}_j$ and $o^{i-1}_j$, i.e., $o^{i}_j-o^{i-1}_j$.

\begin{figure}[h]
\label{assum}
\centering
\includegraphics[width=0.6\linewidth]{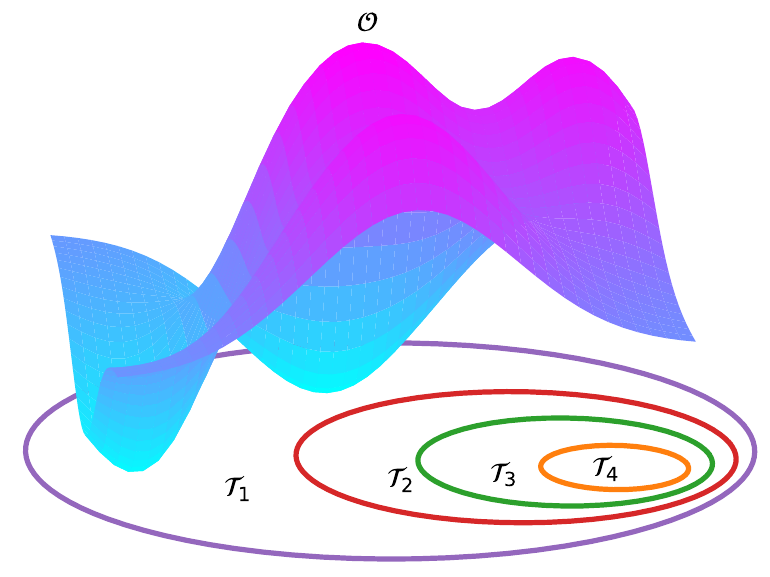}
\caption{\label{Fig1} Distribution discrepancy of different task spaces in SDMTL. The curved surface represents the distribution $\mathcal{O}$ with domain $\mathbb{X}\times\mathbb{T}_1\times\cdots\times\mathbb{T}_N$. The colored circles from the outside to the inside denote the domains of tasks $\mathcal{T}_1$, $\mathcal{T}_2$, $\mathcal{T}_3$, and $\mathcal{T}_4$, respectively.}
\end{figure}

Since there exists the dependence relationship between the previous and current tasks, i.e., $\mathcal{T}_1\to\mathcal{T}_2\to\mathcal{T}_N$, the sample space of the current task depends on that of the previous task. In general, the sample space of the previous task contains the sample space of the current one as shown in Fig. \ref{Fig1}, which leads to the data distribution discrepancy between these two sample spaces. 
Consider the general CTR, CVR and CTCVR estimation tasks, i.e., impression$\to$click$\to\cdots\to$conversion. We use the random variables $Y\in\{0,1\}$ and $Z\in\{0,1\}$ to denote the click event and the conversion event, respectively. Then, CTR, CVR and CTCVR with feature input $X$ are defined as $P(Y\vert{X})$, $P(Z\vert{Y=1},X)$ and $P(Z,Y=1\vert{X})$, which satisfy
\begin{align}
\label{Eq3-2}
P(Z=1,Y=1\vert{X})=P(Y=1\vert{X})P(Z=1\vert{Y=1},X).
\end{align}
In this case, the training space of the traditional CVR estimation task is generally determined by the samples with $Y=1$ in the CTR estimation task. However, for a new user, there are no impression and click records. The conversion rate estimation task is actually to estimate $P(Z=1\vert{X})$. Considering $P(Z=1,Y=0\vert{X})=0$ \cite{o2021analysis} and according to (\ref{Eq2.1_01}), we can obtain 
\begin{align}
\label{Eq3-3}
P(Z=1\vert{X})=&\;P(Z=1,Y=1\vert{X}),\nonumber\\
P(Z=0\vert{X})=&\;P(Z=0,Y=0\vert{X})+P(Z=0,Y=1\vert{X}).
\end{align}
According to (\ref{Eq3-3}), if the negative data points of the event $Z$ only derived from the space with $Y=1$ is used to predict $P(Z=1\vert{X})$, then the data distribution discrepancy between training space and inference space leads to inaccurate predictions.

\subsection{Distribution Dependence Relationship Between Inference Space and Local Space}
In this subsection, the relationship of expected losses between domains of adjacent tasks $\mathcal{T}_{i-1}$ and $\mathcal{T}_i$ is established. 
In tasks $\mathcal{T}_{i-1}$ and $\mathcal{T}_i$, the sample space with data points $\{x_j\in\mathbb{X},o_j^{i-1}\in\{0,1\},o_j^{i}\in\{0,1\}\}$ is called inference space, i.e., entire space for $\mathcal{T}_{i-1}$ and $\mathcal{T}_{i}$, and the sample space with data points $\{x_j\in\mathbb{X},o_j^{i-1}\in\{1\},o_j^{i}\in\{0,1\}\}$ is called local space, also called training space in some traditional CVR estimation methods \cite{ESMM2018}. The distributions of inference and local spaces are denoted as $\mathcal{D}$ and $\mathcal{C}$, respectively.

Therefore, the objective of these two tasks $\mathcal{T}_{i-1}$ and $\mathcal{T}_i$ with sequential dependence in inference space is to minimize the following expected loss:
\begin{align}
\label{Eq3-5}
E&_{X,T_{i-1},T_{i}\thicksim\mathcal{D}}[L(f_{i-1}(X), T_{i-1})+L(f_{i}(X),T_{i})]\nonumber\\
=&E_{X,T_{i-1}\thicksim\mathcal{D}}[L(f_{i-1}(X),T_{i-1})]+E_{X,T_{i}\thicksim\mathcal{D}}[L(f_i(X), T_{i})]\nonumber\\
=&\int_{\mathcal{D}}L(f_{i-1}(x), t_{i-1})P_\mathcal{D}(x,t_{i-1}){\rm d}x{\rm d}t_{i-1}\nonumber\\
&+\int_{\mathcal{D}}L(f_i(x), t_i)P_\mathcal{D}(x,t_i){\rm d}x{\rm d}t_i,
\end{align}
where $P_\mathcal{D}(\cdot,\cdot)$ is the joint distribution in inference space. On the other hand, if the model is trained in the local space $\mathcal{C}$, then the task $\mathcal{T}_i$ determines the sample distribution of the task $\mathcal{T}_{i-1}$. With this operation, the expected loss becomes the following form: 
\begin{align}
\label{Eq3-6}
E_{X,T_{i-1}\thicksim\mathcal{D}}&[L(f_{i-1}(X),T_{i-1})]+E_{X,T_i\thicksim\mathcal{C}}[L(f_i(X), T_{i})]\nonumber\\
=&\int_{\mathcal{D}}L(f_{i-1}(x), t_{i-1})P_\mathcal{D}(x,t_{i-1}){\rm d}x{\rm d}t_{i-1}\nonumber\\
&+\int_{\mathcal{C}}L(f_{i}(x), t_{i})P_\mathcal{C}(x,t_{i}){\rm d}x{\rm d}t_i,
\end{align}
where $P_\mathcal{C}(\cdot,\cdot)$ is the joint distribution in local space. Next, the relationship between expected losses in (\ref{Eq3-5}) and (\ref{Eq3-6}) is revealed.
\begin{theorem}
\label{Th-01}
If the expected losses in the inference and local spaces are defined as in (\ref{Eq3-5}) and (\ref{Eq3-6}),
then, for any loss function $L(\cdot,\cdot)$, they satisfy
\begin{align}
\label{Eq3-7}
E&_{X,T_{i-1},T_i\thicksim\mathcal{D}}[L(f_{i-1}(X), T_{i-1})+L(f_{i}(X),T_{i})]\nonumber\\
=&E_{X,T_{i-1}\thicksim\mathcal{D}}[L(f_{i-1}(X),T_{i-1})]\nonumber\\
&+E_{X,T_{i}\thicksim\mathcal{C}}\Big[P_{\mathcal{D}}(T_{i-1}=1)\frac{P_{\mathcal{D}}(T_i\vert{X})}{P_{\mathcal{D}}(T_i,T_{i-1}=1\vert{X})}L(f_i(X),T_i)\Big].
\end{align}
\end{theorem}

Obviously, the distribution shift also exists in CTR, CVR and CTCVR estimations when they are trained in different spaces. 
\section{The Adaptive Pattern Extraction Multi-task Framework}
\begin{figure}[htbp]
\centering
	\includegraphics[width=0.95\linewidth]{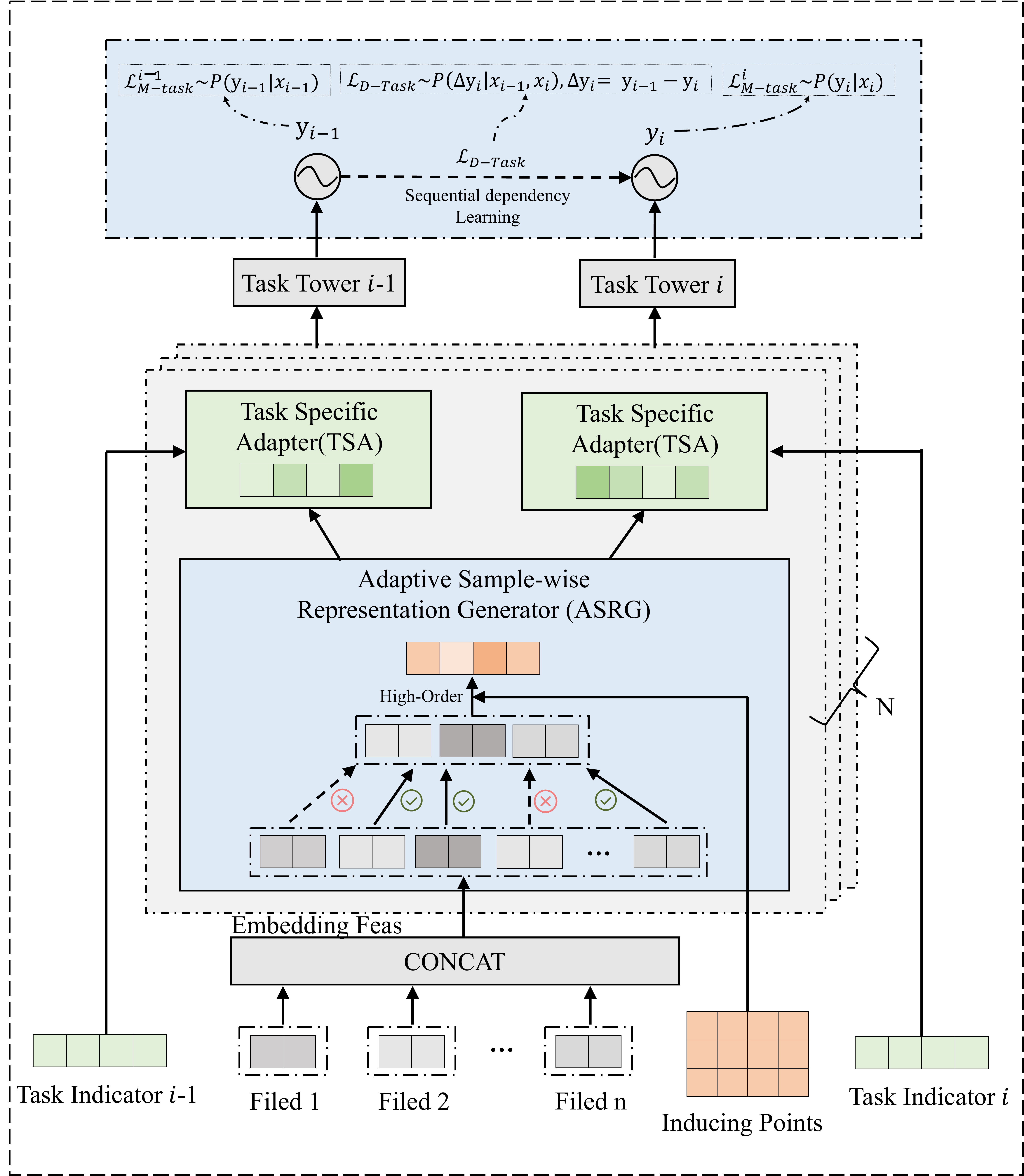}
	\caption{An illustration of the overall architecture of APEM.}
	\label{detail}
\end{figure}

The whole architecture of proposed APEM for sequential dependence multi-task learning is illustrated in Figure~\ref{detail}. APEM consists of two representation learning modules ASRG and PS to dynamically extract implicit and explicit feature information from a sample-wise view, and a sequential dependence task learning loss to reconstruct an unbiased task relationship on a global training space. Adaptive Sample-wise Representation Generator (ASRG) is responsible for hierarchical shared-representation learning, adopting inducing points to interact with different feature field corresponding to each input. Task Specific Adapter (PS) module cooperates with ASRG but works as a task-ware information extractor through designed task indicator and is with an independent message passing structure to better solve the task conflict. Besides those two, a sequence dependency learning loss between tasks is proposed and theoretical proved, which is able to describe the conditional dependent probability for sequential based multi-task learning from the whole training space and consequently improve the prediction result by precisely capturing the task relationship. We will elaborate ASRG and PS in section \ref{asrg_s} and \ref{tsa_s}, and lastly discuss the relationship between sequence dependence tasks in section \ref{loss}.

\subsection{Adaptive Sample-wise Representation Generator}\label{asrg_s}
\begin{figure}[htbp]
\centering
	\includegraphics[width=0.95\linewidth]{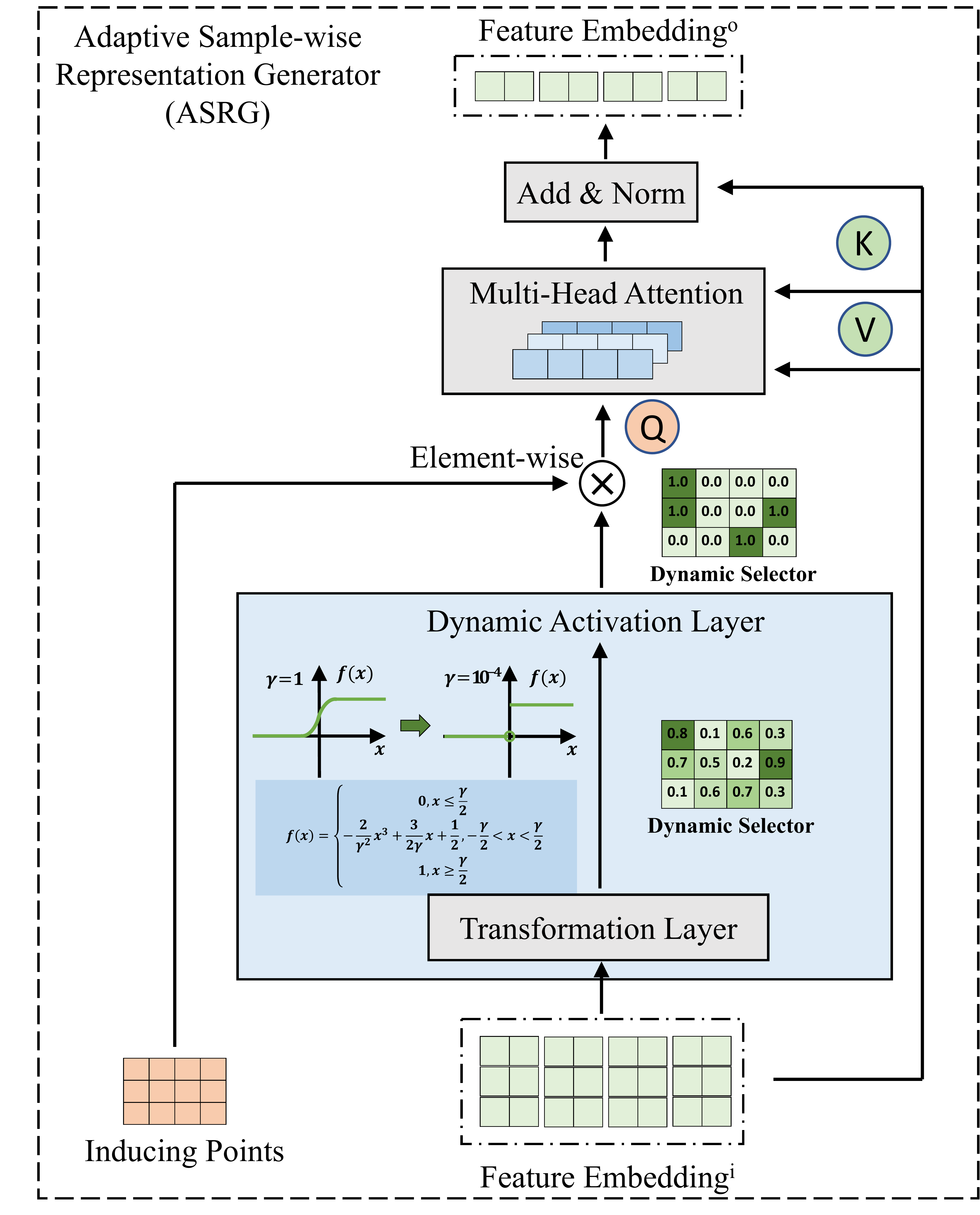}
	\caption{The detail structure of Adaptive Sample-wise Representation Generator}
	\label{ASRG}
\end{figure}

Fine-grained feature information extraction corresponding to different tasks is crucial in multi-task learning and significantly affects model performance. But feature generalization also needs to be included to balance the trade-off between tasks in terms of shared information. Based on these considerations, we propose a novel representation learning module, named Adaptive Sample-wise Representation Generator (ASRG). Besides learning generalized shared-information, we design a dynamic selector to learn the feature interaction from a sample-wise view to further separates the task-irrelevant info. The structure of ASRG is shown in
Figure~\ref{ASRG}, which mainly consists of an dynamic activation layer and a feature interaction learning layer. 

~\\
\textbf{Dynamic Activation Layer}. In recommendation scenario, input field usually contains kinds of user and item features. Given an input $\mathbf{x}$ from $F$ different feature fields, we denote $\mathbf{x}$ as the concatenation of all feature fields:

\begin{equation}
	\mathbf{x}=[x_{1},x_{2},\dots,x_{F}],
\end{equation} 
where $x_i$ represents the value of the $i$-th feature. As a commonly data preprocssing for online recommendation scenario with better generalization, we  discretize numerical features $x_i$ through a Log-round operation to get an unique value, and randomly initialize it with a vector of $d_{f}$ dimension. Thus, we obtain the input embedding for each feature field as $H=[{h}_{1},{h}_{2}, \dots, {h}_{F}]^{\mathrm{T}}$, where $H \in \mathbb{R}^{F\times d_{f}}$. 

\begin{figure}[htbp]
\centering
	\includegraphics[width=1\linewidth]{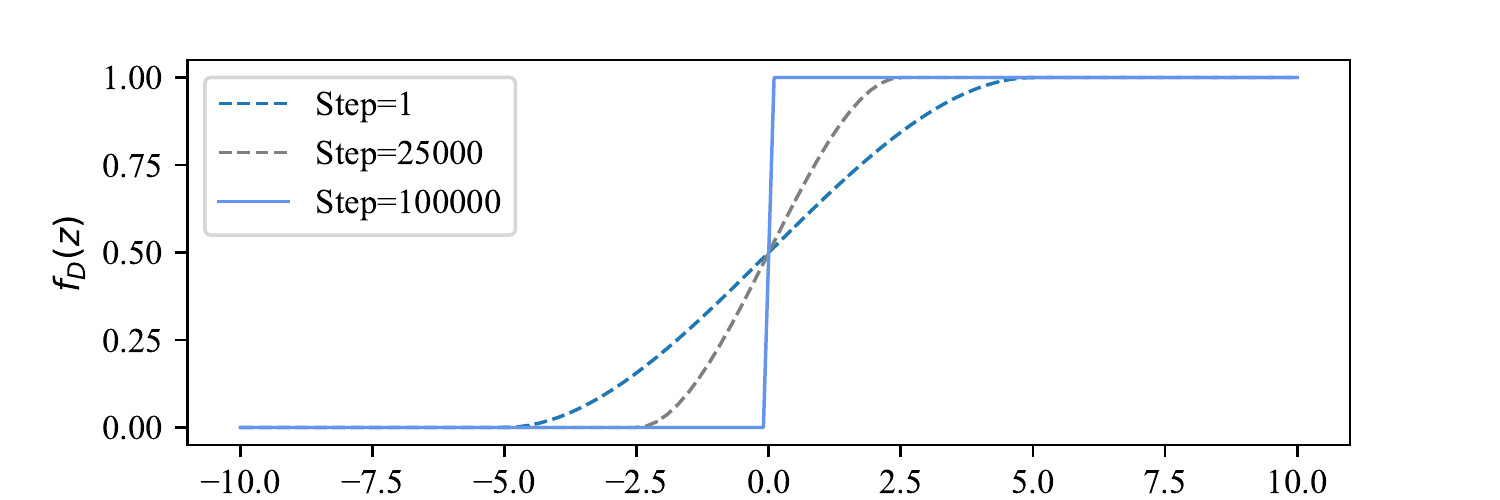}
	\caption{Output of the dynamic activation function $f_{D}$ with the increase of training step.}
	\label{gamma}
\end{figure}

A transformation Layer is first applied to project the input embeddings into a $K$ dimension vector. The transformation layer can be any type of deep neural network structure and here we chose a standard MLP layer just for simplicity. The output $z_K$ is defined as a dynamic selector:
\begin{equation}
	z_K =\operatorname{MLP}(H)
\end{equation} 
where $z_K \in \mathbb{R}^{K}$. Then, we implement a dynamic activation function $f_{D}$ inspired by ~\cite{2021DSelect} to get a sparser representation of $z_K$, whose formulation is as follows:
\begin{equation}\label{dynamic seletor}
	z_K =f_{D}(z_K)
\end{equation} 
where $f_{D}$ is formulated as:
\begin{equation}
\label{dynamic}
f_{D}(z)=\left\{
\begin{array}{lc}
0, &z \leq -\frac{\gamma}{2} \\
-\frac{2}{\gamma^3}z^3 + \frac{3}{2\gamma}z + \frac{1}{2}, &-\frac{\gamma}{2} < z < \frac{\gamma}{2} \\
1, &z \geq \frac{\gamma}{2}
\end{array}
\right.
\end{equation}
where $\gamma=Max\{10 - 2e\textnormal{-}4\cdot step, 1e\textnormal{-}3\}$ and maximum $step$ during the training process is around $1e6$. Dynamic selector $z_K$ works as a information filter which selectively interacts with input from the sample-wise view due to the Transformation Layer. As visualized in Figure~\ref{gamma}, the output shape of $f_{D}$ becomes steeper with the increase of training step. By utilizing  $f_{D}$, $z_K$ creates a $K$ dimension sparse vector only contains values of 0 and 1 corresponding to each input sample.

~\\
\textbf{Feature Interaction Learning Layer}. Attention mechanism for learning hierarchical feature interaction is generally adopted but requires quadratic time complexity in standard self-attention structure. Here we design a learnable matrix called inducing points $I$, enlightened from Set Transformer~\cite{lee2019set} to reduce the computational complexity from quadratic to linear. We define the inducing points $I \in \mathbb{R}^{K \times d_{f}}$, where $K$ is the same as in dynamic selector $z_K$. After a element-wise operation, we get a modified query $\hat{Q}$ as:
\begin{equation}
	\hat{Q} =I \odot z_{K}
\end{equation} 
Then, we calculate the output $O_j$ from the attention operation according to the following formulation:
\begin{equation}
\label{attention}
	O_j = \operatorname{Attention}(\hat{Q}_j, K_j, V_j;\lambda)
\end{equation} 
where $\hat{Q}_j = I_j \odot z_{K}, K_j = HW^K_j, V_j = HW^V_j$ with trainable parameter $\lambda = \left\{I_j, W_j^K, W_j^V\right\}_{j=1}^m$ and $m$ represents the number of multi-head. Then we get output $O$ from the multi-head attention with parameter $W^O$ as:
\begin{equation}
	O = \operatorname{concat}\left(O_{1}, \dots, O_{h}\right)W^{O}
\end{equation} 
Finally, the adaptive representation $Y_{ASRG}$ learned from ASRG can be formulated in a way of residual network:
\begin{equation}
\label{layernorm}
	Y_{ASRG} = \operatorname{LayerNorm}(O, H)
\end{equation} 

The time complexity of feature interaction learning layer reduces from $O(F^2)$ to $O(K\times F)$ by introducing $I$. As suggested in ~\cite{pritzel2017neural}, $K$, the reduced dimension of $I$ could be viewed as $K$ independent memory cells interacting with each feature field, which is further automatically selected by $z_k$ to distinguish feature information explicitly from a sample-wise view. Compared with traditional shared-representation learning structure in most MTL methods, ASRG learns more distinctive info in terms of a dynamic activation layer and feature interaction learning layer, which is mainly attributed to the former one combining a transformation layer and dynamic activation function to generate an adaptive mask corresponding to each input sample.

\subsection{Explicit Pattern Selector} \label{tsa_s}
Besides effective shared-representation generated by ASRG, specific feature learning according to each task will strongly affect the model performance since it directly enhances the task-relevant information. In most MTL works, task-targeted feature extractors, such as the task-specific experts proposed by PLE are deliberately designed to learn the representation for each task. However, mutual interference between different tasks still exists since the shared and task-specific components are not completely separated in these cases.

\begin{figure}[htbp]
\centering
	\includegraphics[width=0.95\linewidth]{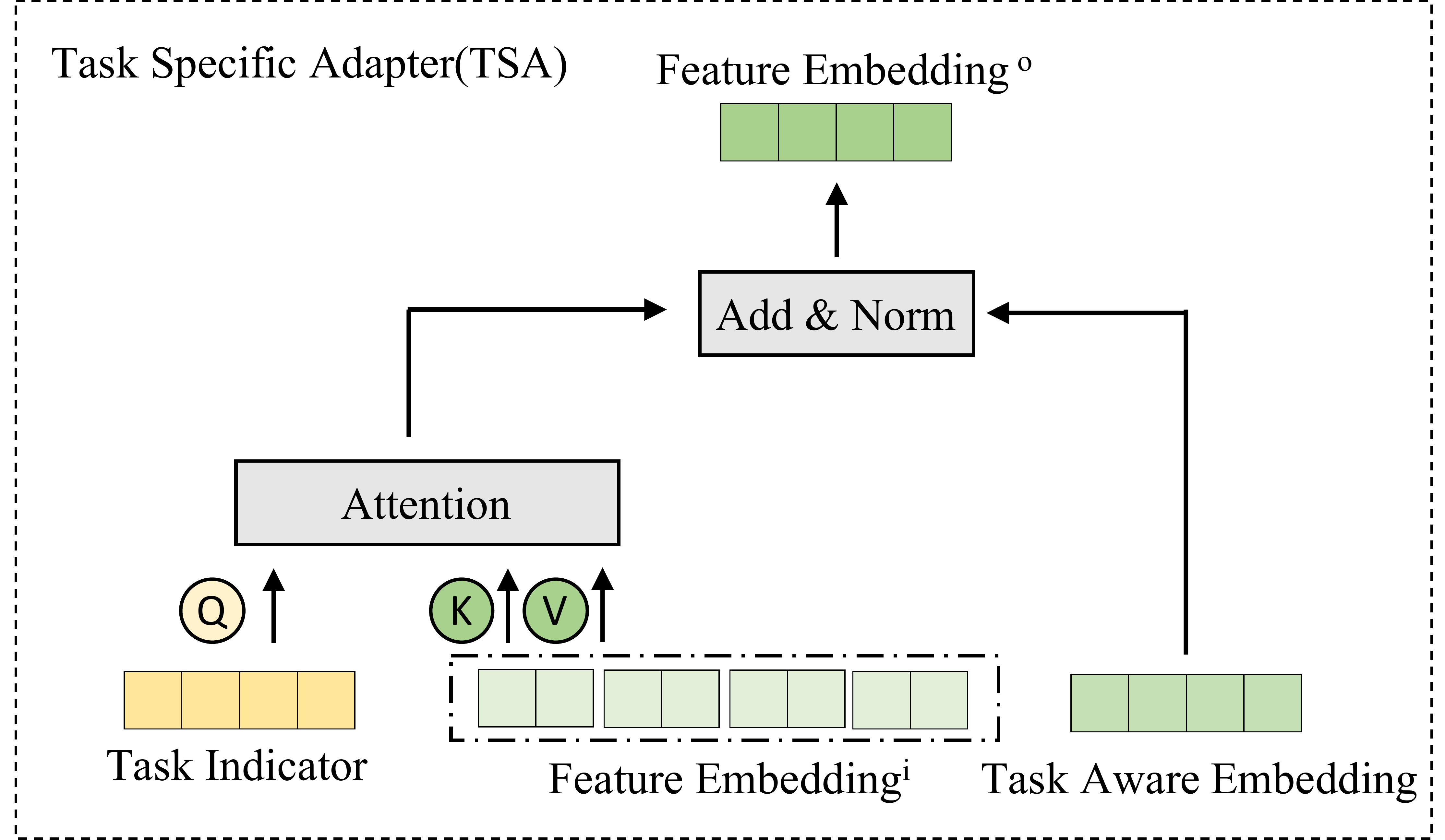}
	\caption{The detail structure of Pattern Selector.}
	\label{tsa}
\end{figure}

In order to learn the task-aware information among different tasks with a more independent and thoroughly separated structure, we introduce a module named explicit Task Specific Adapters (PSs) as detail plotted in Figure~\ref{tsa}. PS utilizes parameterized task indicator vector to interact with previous sample-wise common shared info from ASRG, which is able to extract task-specific representation by directly optimizing respective task object. The approach is similarly adopted in PAL~\cite{stickland2019bert} and K-adapter~\cite{wang2020k}. 

As illustrated in Figure~\ref{tsa}, we take the output $Y_{ASRG} \in \mathbb{R}^{K\times d_f}$ from the Adaptive Sample-wise Representation Generator to interact with a learnable task indicator vector $\alpha_i$ corresponding to each task $i$. The $F_i$ is the output calculated through an attention operation between $Y_{ASRG}$ and $\alpha_i$ as:
\begin{equation}
	F_i = \operatorname{Attention}(\alpha_i, Y_{ASRG}, Y_{ASRG}),
\end{equation}
where $\alpha_i \in \mathbb{R}^{1 \times d_f}$ is the task indicator vector 
and $F_i \in \mathbb{R}^{1 \times d_f}$ denotes the middle output of task-aware representation for each task $i$ correspondingly. Here, $\operatorname{Attention}$ is the same attention calculation operation as in formula (\ref{attention}). Consequently, for each task $i$, we refer the task-specific information generated from the $k$-th PS layer as $T_i^k$ and we calculate it also through a residual network and layer normalization for training efficiency as in (\ref{layernorm}):
\begin{equation}
	T_i^k = \operatorname{LayerNorm}(T_i^{k-1} + F_i^k),
\end{equation}
where $T_i^{k-1} \in \mathbb{R}^{1 \times d_f}$ means the output from the previous PS layer for task $i$, and $F_i^{k} \in \mathbb{R}^{1 \times d_f}$ is the task-aware embedding learned by the interaction between the task indicator for the $k$-th layer with task $i$ and shared-common embedding. Note, $T_i^0$ is ignored at the first iteration.

It can be observed in Figure~\ref{tsa}, task aware embedding $T$ obtained from Task Specific Adapters is trained independently and whose message doesn't pass into ASRG module among different layers. The proposed structure keeps the implicit (from ASRG) and explicit (from PS) representation learning modules more separated, which not only isolates the negative interference between tasks more thoroughly but also provides a extendable multi-task learning framework especially necessary in industrial implementation.

\subsection{Loss Function Design Towards Sequential Dependence Multi-Task Learning}\label{loss}
For the multi-task learning without sequential dependence, the loss function is generally designed as the following form:
\begin{align}
\label{Eq4-n-1}\mathcal{L}(\theta^s,\theta^{1},\dotsc,\theta^{N})=\sum_{i=1}^{N}\sum_{j=1}^{M}\frac{w_i}{M}L\Big(f_i(x_j;\theta^s,\theta^i),o_j^i\Big).
\end{align}
From the loss function (\ref{Eq4-n-1}), it can be observed that this loss function cannot learn the sequential dependence relationship. 
As mentioned in subsection \ref{ProblemFormulation}, in this paper, the constrained optimization problem can be transformed into the unconstrained case by using a penalty function. Therefore, 
the corresponding loss function for SDMTL is designed as 
\begin{align}
\label{Eq4-n-4}
&\mathcal{L}(\theta^s,\theta^{1},\dotsc,\theta^{N})\nonumber\\
&=\mathcal{L}_{M-Task} + \mathcal{L}_{D-Task}\nonumber\\
&=\sum_{i=1}^{N}\frac{w_i}{M}\sum_{j=1}^{M}L\Big(f_i(x_j;\theta^s,\theta^i),o_j^i\Big)\nonumber\\
&+\sum_{i=2}^{N}\frac{\sigma_{i-1}}{M}\sum_{j=1}^{M}L\Big(f_{i-1}(x_j;\theta^s,\theta^{i-1})-f_i(x_j;\theta^s,\theta^i),o^{i-1}_j-o^{i}_j\Big),
\end{align}
where $\sigma_{i}$ is the penalty coefficients, $\mathcal{L}_{M-Task}$ and $\mathcal{L}_{D-Task}$ are the loss functions of the main tasks, i.e., $\mathcal{T}_i$, and the loss functions of the sequential dependence relationship, respectively. With this operation, each task and their corresponding dependence relationship can be trained separately. The loss functions of the dependence relationship $\mathcal{L}_{D-Task}$ can be regarded as a regularization term.
Note that the selection of negative samples determines the training space. Therefore, the positive samples of the task $\mathcal{T}_i$ are derived from the current task while the negative samples of $\mathcal{T}_i$ are derived from different tasks $\mathcal{T}_{i}$ and $\mathcal{T}_{i-1}$. Similar to the subsection 3.2, expected losses of the dependence relationship derived from the entire space $\mathcal{D}$ and local space $\mathcal{C}$ are discussed as follows.
\begin{theorem}
\label{Th-02}If the dependence relationship is learned in the entire space $\mathcal{D}$ and the local space $\mathcal{C}$, respectively, and the corresponding expected losses are denoted as $E_{X,T_{i-1},T_i\thicksim\mathcal{D}}[L(f_{i-1}(X)-f_i(X),T_{i-1}-T_i)]$ and $E_{X,T_{i-1},T_i\thicksim\mathcal{C}}[L(f_{i-1}(X)-f_i(X),T_{i-1}-T_i)]$, then these two expected losses satisfy
 \begin{align}
\label{Eq4-n-5}
E_{X,T_{i-1},T_i\thicksim\mathcal{D}}&[L(f_{i-1}(X)-f_i(X),T_{i-1}-T_i)]\nonumber\\
=&\;E_{X,T_{i-1},T_i\thicksim\mathcal{C}}\Bigg[\frac{P_{\mathcal{D}}(T_{i-1}=1)P_{\mathcal{D}}(T_{i-1}-T_i\vert{X})}{P_{\mathcal{D}}(T_{i-1}-T_i,T_{i-1}=1\vert{X})}\nonumber\\
&\times{L}(f_{i-1}(X)-f_i(X),T_{i-1}-T_i)\Bigg]
\end{align}
\end{theorem}

\section{Experiments}
In this section, we describe the experiments to evaluate the  performance of the proposed APEM framework, which are conducted on both public benchmark dataset and real-world industrial dataset in financial service. We also analyze the contribution of each modules consisting of APEM to further understand the working mechanism and demonstrate the effectiveness of proposed method for sequential dependence multi-task learning.
\subsection{Experimental Setup}
\subsubsection{Datasets}
Experiments are conducted on two dataset: the public benchmark Ali-CCP and an industrial dataset from the financial scenario.
\begin{itemize}
    \item \textbf{Ali-CCP dataset:} \footnote{\url{https://tianchi.aliyun.com/dataset/dataDetail?dataId=408}} The public dataset Ali-CPP is used as the benchmark for model comparison with Alibaba Click and Conversion Prediction tasks. We use all the single-valued categorical features as generally adopted and randomly take 10$\%$ of the train dataset as the validation dataset for all models.
    \item \textbf{Industrial Dataset:} The industrial dataset is collected from our online recommendation platform in financial service, which describes users' clicking and conversion behavior responding to financial advertising. The dataset is divided for training, validation and test chronologically and we downsample the negative samples in the training set to keep the ratio of each set is 8:1:1.   
\end{itemize}

\subsubsection{Baseline Methods}
To validate the effectiveness of APEM, we conduct our experiments on the following representative methods for comparison, which are SOTA multi-task learning approaches or recent sequence dependency learning method:
\begin{itemize}
    \item \textbf{Single-Task} is a three-layer MLP network with hidden layer size of [256,128,64] for single-task optimization.
    \item \textbf{Shared-Bottom} constructs a shared bottom layer to learn the common representation across all tasks and introduces separated task tower for the object optimization respectively.
    \item \textbf{MMOE} is inspired by the classic MoE method which adopts a group of shared bottom subnetworks as experts and introduces gating network assigning different tasks with distinctive weights.  
    \item \textbf{PLE} generalizes CGC method and employs a progressive routing mechanism to extract and separate deeper semantic knowledge.
    \item \textbf{AITM} is a shared-bottom structure with adaptive information transfer for modeling sequential dependency among multi-step conversions.
    \item \textbf{APEM} is our proposed approach which adopts adaptive sample-wise representation generator and the explicit Pattern Selector, with dependency learning loss for the sequence dependence multi-task learning.
\end{itemize}

\subsubsection{Implementation of $\mathcal{L}_{D-Task}$}
As discussed in section \ref{loss}, $\mathcal{L}_{D-Task}$ can be regarded as a regularization, which constraints the relationship between sequential dependence tasks during training process. In this paper, we constructs a MSE (Mean-Squared Loss) as an implementation for $L$ in formula (\ref{Eq4-n-4}):

\begin{equation} \label{l-d}
	L_{MSE} = \frac{1}{M}\sum_{j=1}^{M}w\cdot(y_j-\hat{y}_j)^2
\end{equation}
where $y_j$ is label from $o_j^{i-1}-o_j^i$ and $\hat{y_j}$ is the output from $f_{i-1}(x_j;\theta^s,\theta^{i-1})-f_i(x_j;\theta^s,\theta^i)$ for input $j$. Each sample is equally treated with $w=1$.

\subsubsection{Training Setup}
In the experiments, each experiment is repeated 5 times, the average performance and the p value are both reported. We select the optimal hyper-parameters for each model in terms of grid search~\cite{lerman1980fitting} for fair comparison. The batch size $B$ on each datasets is set as $1024$ respectively during the training process.  Adam optimizer~\cite{kingma2014adam} is applied with a learning rate $\lambda$ of 0.001. The dimension $d_f$ of input embedding layer is  $18$. The number of the stacked layers $L$, the number of the attention heads $M$ and the number of the inducing points $K$ is illustrated in Table \ref{training setup}. The activation function of MLP in single-task modeling is ReLU.\\

\begin{table}[]

\footnotesize
    \captionsetup{}
    \caption{Detailed hyper-parameters settings for each dataset.}
    \begin{center}
    \begin{tabular}{c|l}
        \hline
        {Dataset} & {Hyper-parameters Settings}  \\
        \hline
        {Ali-CCP} & $B=1024,d_f=18,M=2,K=64,L=4,\lambda=10^{-3}$ \\ 
        {Industrial dataset} & $B=1024,d_f=18,M=2,K=64,L=4,\lambda=10^{-3}$\\ 
        \hline
    \end{tabular}
    \end{center}
    \label{training setup}
\end{table}



\subsection{Performance Comparison}\label{Performance Comparison}

\begin{table*}[]
\scriptsize
\caption{The performance (AUC) comparison with baselines.The Gain means the mean AUC improvement compared with Single-Task method. ** indicates that the improvement of the proposed APEM is statistically significant compared with the best baseline at a p-value < 0.01 over paired samples t-test.}
\label{tab1}
\begin{center}
\setlength{\tabcolsep}{5mm}{
\begin{tabular}{ccccc|cccc}
\toprule
Models & \multicolumn{4}{c}{Ali-CPP} & \multicolumn{4}{c}{Industrial Dataset} \\ 
& CTR & CVR & $Gain_{CTR}$ & $Gain_{CVR}$   & CTR  & CVR  & $Gain_{CTR}$ & $Gain_{CVR}$   \\
\midrule
Single-Task  &0.6089 & 0.6011 & -- & -- & 0.7081 & 0.7616& -- &  -- \\ 
Shared-Bottom & 0.6098 & 0.6225 & 0.15$\%$ &  3.56$\%$ & 0.7050&	0.7614& -0.37$\%$ &  0.11$\%$      \\ 
MMOE &0.6177 &0.6223 &1.45$\%$& 3.53$\%$ & 0.7134&	0.7673& 0.82$\%$ &  0.90$\%$   \\ 
PLE &\underline{0.6195} &0.6355 &\underline{1.74$\%$}&5.72$\%$ & 0.7140&	0.7675& 0.90$\%$ &  
0.92$\%$     \\
AITM &0.6133 &\underline{0.6391}&0.72$\%$&\underline{6.32$\%$} & 0.7110&	0.7667& 0.47$\%$ & 0.81$\%$     \\ 
ESMM &0.6193 &0.6333 &1.71$\%$&5.36$\%$ & 0.7154&	0.7680& 1.10$\%$ & 0.99$\%$     \\ 
ESCM2 &0.6153 &0.6258 &1.05$\%$&4.11$\%$ & \underline{0.7146}&	\underline{0.7701}& \underline{0.99$\%$} & \underline{1.26$\%$}     \\ 
\midrule
APEM &\textbf{0.6198} &\textbf{0.6436**}&\textbf{1.79$\%$}&\textbf{7.07}$\%$ &\textbf{0.7167**}&	\textbf{0.7714**} & \textbf{1.29$\%$} &  \textbf{1.43$\%$}    \\
\bottomrule
\end{tabular}}
\end{center}
\end{table*}


\begin{table*}[]
\scriptsize
\caption{The performance (AUC) comparison of ablation study.}
\label{tab1}
\begin{center}
\setlength{\tabcolsep}{5mm}{
\begin{tabular}{ccccc|cccc}
\toprule
Models & \multicolumn{4}{c}{Ali-CPP} & \multicolumn{4}{c}{Industrial Dataset} \\ 
& CTR & CVR & $Gain_{CTR}$ & $Gain_{CVR}$   & CTR  & CVR  & $Gain_{CTR}$ & $Gain_{CVR}$   \\
\midrule
APEM w/o ASRG  &0.6178  & 0.6379 & -0.32$\%$ & -0.89$\%$  & 0.7131 & 0.7670& \textbf{-0.50$\%$} & \textbf{-0.57$\%$} \\ 
APEM w/o PS &0.6158 & 0.6382 & \textbf{-0.65$\%$} &  -0.84$\%$ & 0.7141 & 0.7695& -0.36$\%$ &  -0.25$\%$      \\ 
APEM w/o $\mathcal{L}_{D-Task}$ &\textbf{0.6199} & 0.6319 &0.02$\%$& \textbf{-1.82$\%$} & 0.7160 & 0.7695& -0.10$\%$ &  -0.25$\%$   \\ 

\midrule
APEM &0.6198 &\textbf{0.6436}&--&-- &\textbf{0.7167}&	\textbf{0.7714} & -- &  --    \\
\bottomrule
\end{tabular}}
\end{center}
\end{table*}



The experimental results for all comparison methods with the evaluation metric AUC for each task are presented in Table ~\ref{tab1}. 
The best performance on different datasets are highlighted in boldface and underline for the best SOTA mehods. As can be observed, APEM outperforms most baseline models for each task on both datasets respectively.

The average performance on Ali-CCP dataset is poor both on CTR and CVR targets on all compared methods, which probably implies the input features are not qualified enough to express the targets or the irrelevant information affects significantly.
For the latter case, task-specific feature extraction will play a key role to the prediction results in terms of filtering negative interference. As observed, APEM achieves 0.6203 and 0.6456 of AUC for CTR and CVR tasks respectively, with gains of 1.16$\%$ and 7.31$\%$ compared to the Singel-Task method. The improvements of CTR is significant with comparison to other methods but slightly poorer than PLE in the object of CVR. The result seems to be attributed to the trade-off between tasks considered by APEM and APEM. The difference improvements between CTR and CVR is smaller in APEM and suggests a more balanced optimization among tasks.
The performance on the Industrial dataset of APEM obtains an considerable gain of 1.29$\%$ and 1.43$\%$ for both targets and significantly outperforms other methods. Compared with PLE, which achieves a second best result, the proposed model still gets an increase of the gain by 43$\%$ and 55$\%$ and further demonstrates its effectiveness.



\subsection{Ablation Study}
    


We conduct ablation study on different submodules in APEM in order to provide a detail analysis of its function and efficiency. The variant models of APEM consists of following structures and the notation is just for simplicity:
\begin{itemize}
    \item \textbf{APEM without ASRG}: removing the dynamic activation layer in ASRG and replacing with a standard self attention operation.
    \item \textbf{APEM without PS}: removing task indicator in PS layers for all corresponding tasks.
    \item \textbf{APEM without $\mathcal{L}_{D-Task}$}: removing the sequence dependence learning loss $\mathcal{L}_{D-Task}$ as denoted as in (\ref{Eq4-n-4}).
    \item \textbf{APEM}: complete structure of APEM.
\end{itemize}
The results of ablation study are presented in Table ~\ref{abalationtable} with an evaluation metric AUC on both datasets for CTR and CVR tasks. As observed, the complete structure of APEM outperforms all other APEM-variants and we can draw the following conclusions for each submodule:\\
\textbf{(1)}. Adaptive Sample-wise Representation Generator contributes to learn fine-grained and generalized shared representation for both tasks. In which, dynamic selector enables to select essential information for each sample which enhance the knowledge learning. Without the dynamic selection layer, model performance drops most for both CTR and CVR target as -0.5$\%$ and -0.57$\%$ in Industrial dataset. Fully interaction learning via a standard multi-head self-attention can't provide enough shared info. We believe that ASRG reconstruct the necessary information in an adaptive manner which not only learns the feature field interaction but filter the noise by utilizing a group-level attention from a sample-wise view. The contribution in Ali-CPP is still obvious since whose features seems less expressive as discussed in section \ref{Performance Comparison} and is greatly benefited by ASRG. \\
\textbf{(2)}. Explicit Pattern Selectors works as a task-sensitive feature extractor, which is quite crucial in the multi-task learning to precisely extract task-relevant information for each task. It can be evidently observed that without the task attention mechanism (proposed task indicator), model performance drops dramatically and is slightly better than without ASRG in Industrial dataset but worse in Ali-CPP. It is suggested that a vanilla task-specific tower structure doesn't generate enough information during task optimizing process. \\
\textbf{(3)}. The proposed sequence dependence learning loss $\mathcal{L}_{D-Task}$ based on theoretical proof contributes to the model performance in terms of the additional information passing among related tasks. Although it seems less significant compared to other submodules in Industrial dataset but contributes most in the CVR task in Ali-CPP. CVR task probably depends on CTR task heavily and $\mathcal{L}_{D-Task}$ modifies the biased object of original definition, which further optimizes the parameters by recovering their complete probability-dependent relationship.

\subsection{Analysis of Dynamic Selector}
\begin{figure}[htbp]
\centering
	\includegraphics[width=0.95\linewidth]{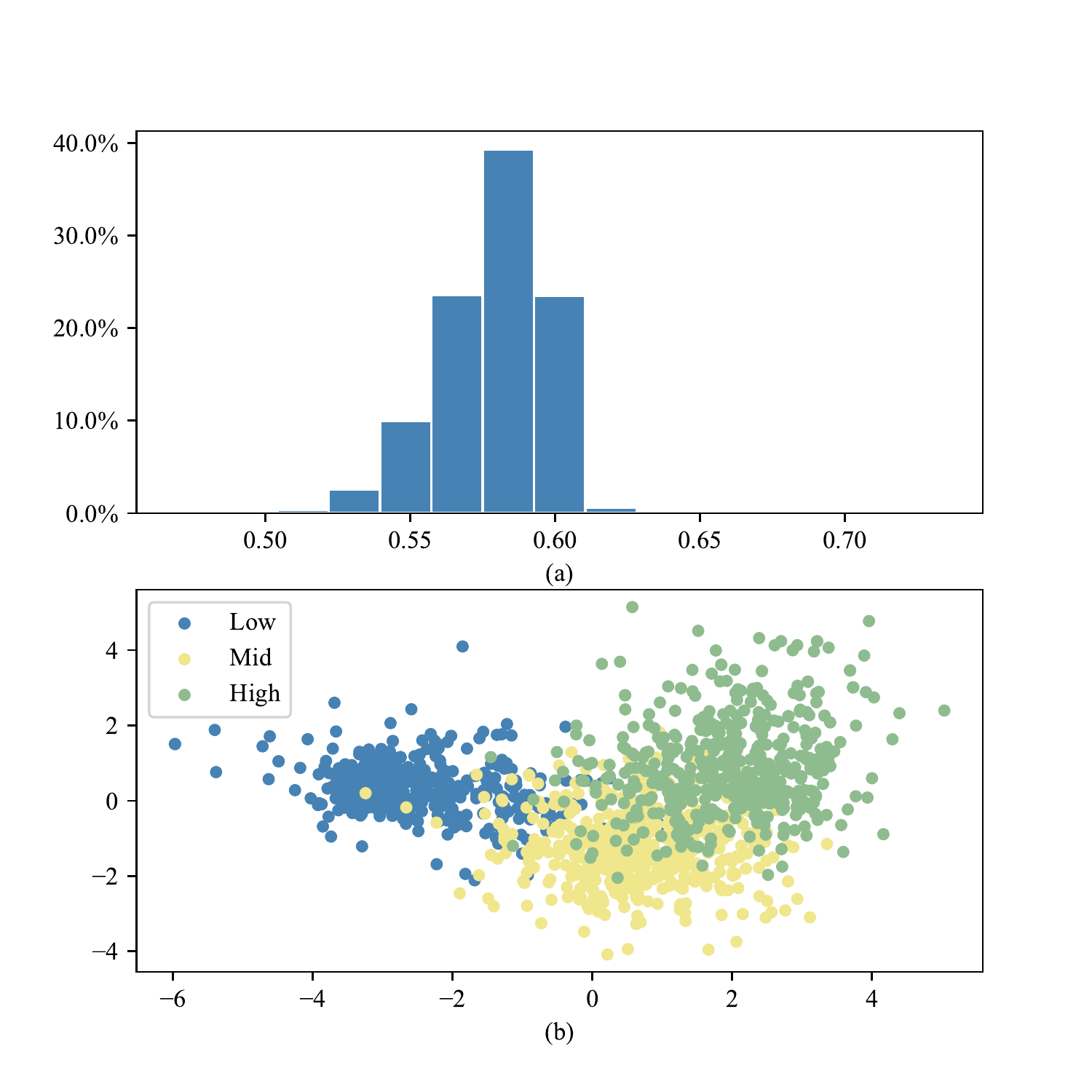}
	\caption{An illustration of dynamic selector $z_K$. (a).Distribution of the selection rate for different samples. (b). Plot of sample embeddings with high, middle and low selection rate is colored in green, yellow and blue respectively.}
	\label{emb_distribute}
\end{figure}

Dynamic selector $z_K$ defined in formula (\ref{dynamic seletor}) functions as a sparse mask generated based on the input sample, which cooperates with Inducing points $I$ interacting with feature fields selectively. We conduct several case studies of $z_K$ to provide an intuitive analysis as visualized in Figure ~\ref{emb_distribute}. 

As noted in section \ref{asrg_s} , $z_K$ is a $K$ dimension vector only with values of 0 and 1, where 1 means interacting with implicit field group in terms of $I$ correspondingly and vice versa. We first plot the distribution of the non-zero rate (selection rate) of $z_K$ on the test samples of industrial dataset in Figure ~\ref{emb_distribute} (a). It can be observed for most samples, the selection rate is between 55$\%$ and 60$\%$ and indicates that more than half the interaction groups are required for information extraction. For specific cases, some needs just less interaction groups and some needs more. We could regard this as a multi-view representations for each sample, such as different numbers of perspectives qualified enough to describe a customer's interest specially on online recommendation scenario. We further randomly plots sample embeddings with high (top 1$\%$), mid (around 58$\%$) and low (bottom 1$\%$) selection rates in Figure ~\ref{emb_distribute} (b). As illustrated, samples with different interaction degrees show significant difference in embedding space and probably implying distinctive intentions. .


\subsection{Efficiency Evaluation}
\begin{figure}[htbp]
\centering
	\includegraphics[width=0.95\linewidth]{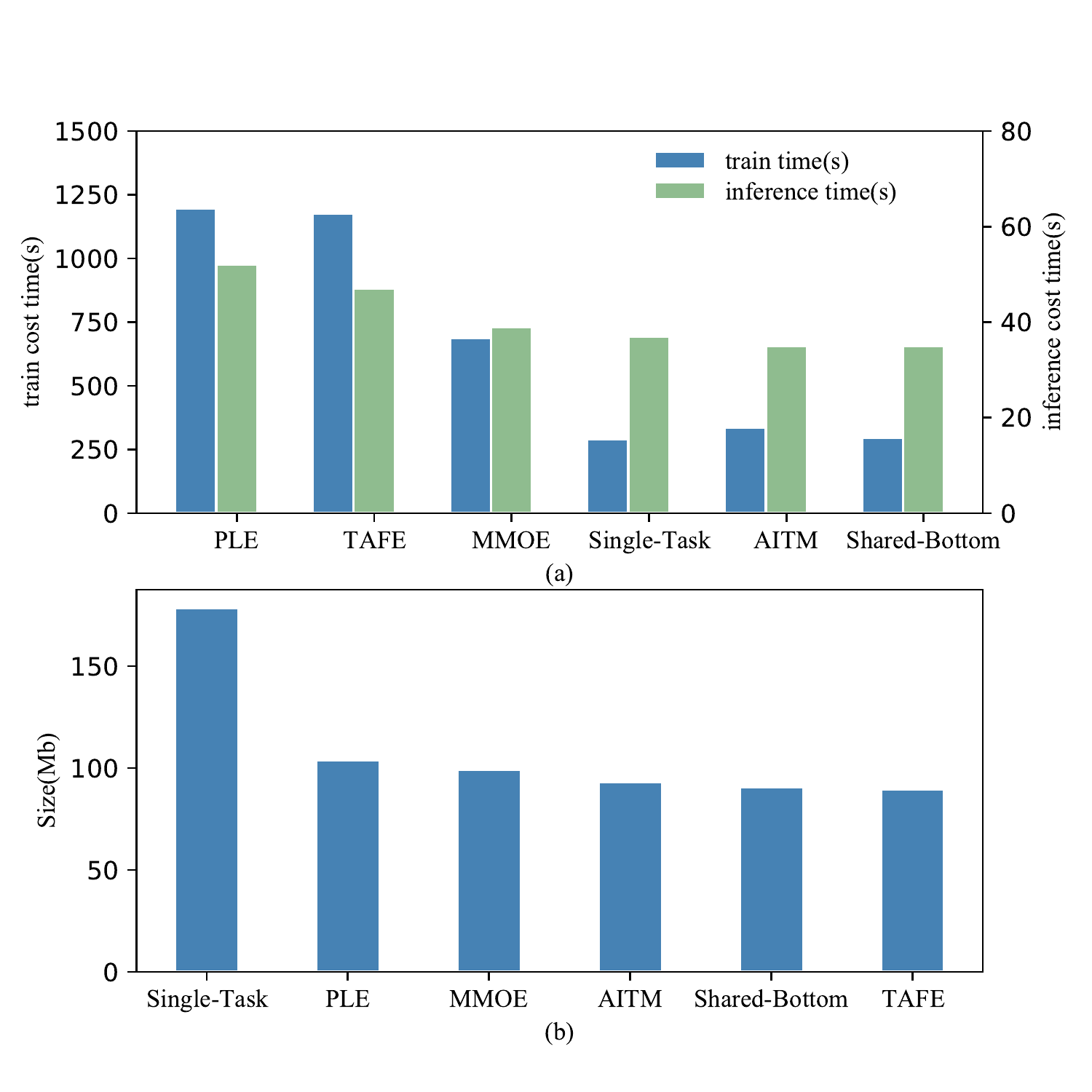}
	\caption{Efficiency comparison among models. (a). Training and Inference time (b). Memory Cost}
	\label{model_size_time}
\end{figure}

In this section, we evaluate time and storage efficiency of our proposed method. We record the time cost during the training (per epoch) and inference process for APEM and other baseline models in Figure ~\ref{model_size_time} (a), and their respective memory cost in Figure ~\ref{model_size_time} (b). As illustrated in (a), APEM requires 1177 seconds for training an epoch on the Ali-CCP dataset with 38 millions samples, which is less efficient than other methods (295s for the best results from Shared-Bottom) but similar to the PLE (1195s). Since we generally train the model in an offline manner especially for large-scale data, relatively higher training efficiency is within tolerance. On the inference time, the essential factor considered in the online industrial application, APEM spends 47 seconds for forward propagation on the test data with 4.2 millions samples. Its deviation from top performance models like AITM and Shared-Bottom is 12 seconds and it is acceptable considering most industrial online inference situation' QPS threshold. Besides, APEM has the least parameters with 89 Mb (178 Mb for the largest model of Single-Task) as plotted in Figure ~\ref{model_size_time} (b), which makes it easily deployed and portable. In a conclusion, APEM achieves an significant improvement with an appropriate computational time and advantageous storage capacity compared with other approved MTL methods.

\subsection{Online A/B Performance}

\begin{table}[]
\centering
\scriptsize
\caption{The offline performances (AUC) comparison for two real-world financial scenes.}
\label{offlinetab}
\begin{center}
\setlength{\tabcolsep}{5mm}{
\begin{tabular}{ccc|cc}
\toprule
Models & \multicolumn{2}{c}{Scene 1} & \multicolumn{2}{c}{Scene 2} \\ 
  & CTR & CVR   & CTR  & CVR   \\
\midrule
MMOE  &0.8102 & 0.8034 & 0.8110 & 0.8719 \\ 
APEM  & 0.8102 & \textbf{0.8072} & \textbf{0.8123} & \textbf{0.8773} \\ 
\midrule
Gain &- &\textbf{0.47$\%$}&\textbf{0.16$\%$}& \textbf{0.62$\%$}\\
\bottomrule
\end{tabular}}
\end{center}
\end{table}

\begin{figure}[htbp]
	\includegraphics[width=0.95\linewidth]{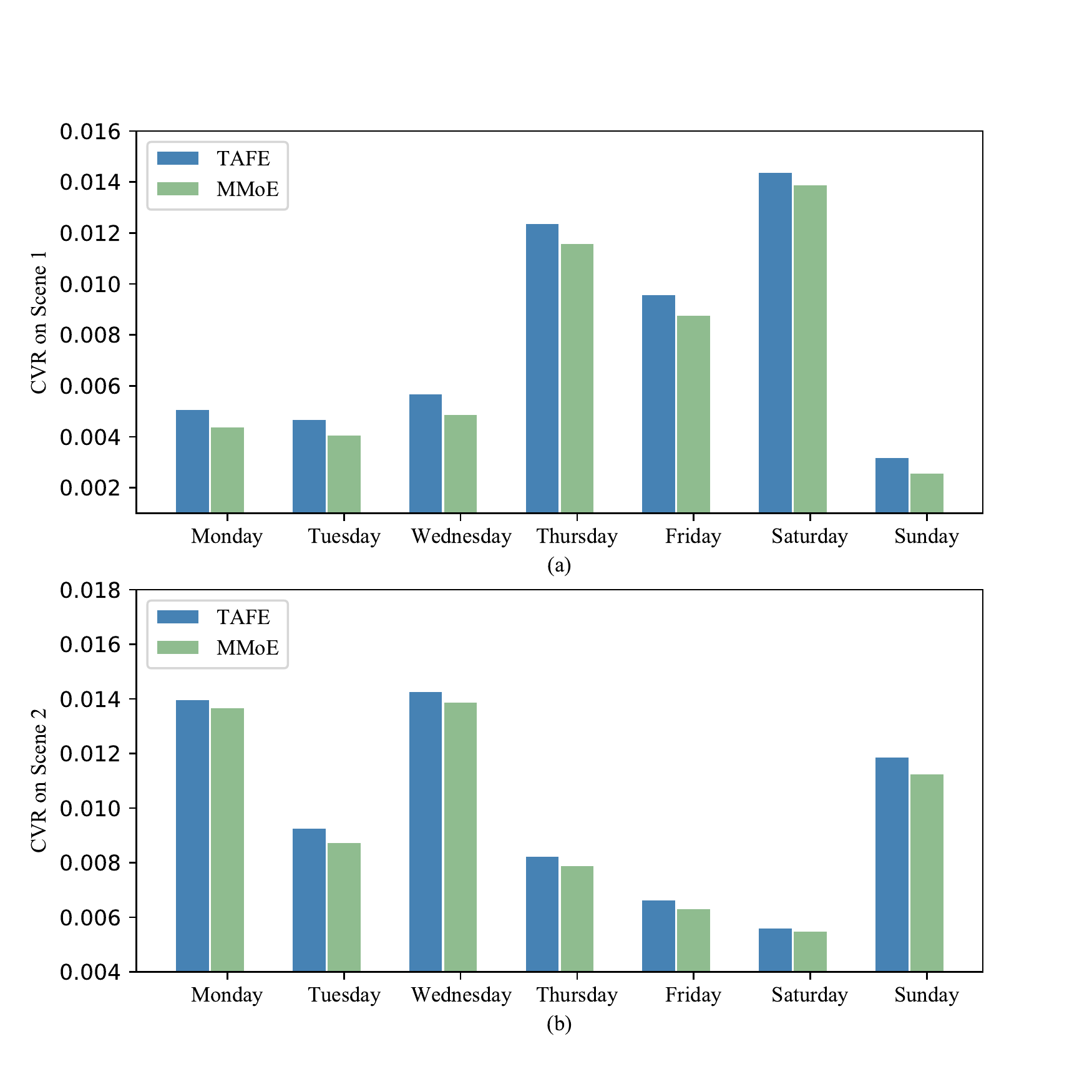}
	\caption{Online A/B results on two real-world scenarios (a) and (b).}
	\label{online_abtest}
\end{figure}

 We implement an online A/B test between APEM and SOTA multi-task learning method of MMOE for one week. Two models are deployed on two real-world financial advertising scenarios with the objective of maximizing the CVR for the financial products as investment and credit loan. The offline comparison is presented in Table ~\ref{offlinetab}, APEM gets an improvement of 0.16$\%$ for CTR in Scene 2, 0.47$\%$ and 0.62$\%$ for CVR on each scene correspondingly. In Figure ~\ref{online_abtest}, we can observe the online performances of APEM compared with MMOE. As shown in the plot, APEM achieves significant and consistent improvements for both scenarios during the whole period, average increase of 9.22 $\%$ in scenario (a) and 3.76$\%$ in scenario (b) on the CVR task. The experiment result proves the efficiency and the stability of proposed mehtod, which is qualified enough for large-scale industrial application.

\section{Related Work}
\textbf{Multi-task Learning.}
Multi-task Learning (MTL) is proposed to learn the shared information among tasks to improve the model generalization and performance ~\cite{caruana1997multitask}. However, multi-task learning scenario usually suffers from the performance deterioration as negative transfer because of the complex relationship between different tasks ~\cite{baxter1997bayesian,vafaeikia2020brief}. Therefore, much feature learning works in structure designing are proposed for necessary information extraction according to specific task and balancing the performances among all tasks. Cross-Stitch Network~\cite{misra2016cross} use a linear combination of shared representations to learn the task-specific embeddings for each task. Based on the idea of Cross-Stitch, Sluice Network~\cite{ruder2019latent} is a generalized meta-architecture with more task-specific parameters by dividing each layer into task-specific and shared subspaces and achieves better performance specially for less correlated tasks. However, these approaches could not capture the sample dependence and require more training data and less efficient for large-scale 
application. Inspired by the MoE~\cite{jacobs1991adaptive} structure, multi-gate Mixture-of-Experts (MMoE) ~\cite{ma2018modeling} employs an ensemble of experts submodules and gating network to combine the representation of the bottom experts to learn the task relationship while consuming less computation. Similarly, Multiple Relational Attention Network (MRAN)~\cite{zhao2019multiple} models multiple relationships by three attention-based learning mechanism. Compared with MMoE, Progressive Layered Extraction (PLE) method~\cite{tang2020progressive} propose a novel MTL framework which separates task-common and task-specific parameters more explicitly and adopts a progressive separation routing mechanism to better alleviate parameter conflicts caused by complex task correlation. 
\\
~\\
\textbf{Sequential Dependence Multi-task Learning.}
The most classical applications of the sequential dependence MTL (SDMTL) are the multi-step conversion process of customer acquisition in e-commerce, display advertising or finance systems. In general, the multi-step conversion process involves impression$\to$click$\to\cdots\to$conversion, which corresponds to several estimation tasks like post-view click-through rate (CTR), post-click conversion rate (CVR) and post-view click-through \& conversion rate (CTCVR) estimations and so fourth. Differing from the general MTL, there exist dependence relationships between the adjacent tasks in the SDMTL problem. For the CVR estimation problem, Entire Space Multi-task Model (ESMM) is proposed in \cite{ESMM2018} to overcome Sample Selection Bias (SSB) and Data Sparsity (DS) issues by introducing two auxiliary tasks of predicting CTR and CTCVR. With this operation, the performance of the CVR estimation will depend heavily on the performance auxiliary tasks. As the number of steps increases in multi-step conversion path, the accumulation of performance errors becomes intolerable. Aimed at the DS problem of the CVR estimation, in \cite{Entire_SIGIR_2020}, a novel user sequential behavior graph is established to achieve post-click behavior decomposition by inserting disjoint purchase-related deterministic action and other action into between click and conversion. Considering micro behaviors (user’s interactions with items) and macro behaviors (user’s interactions with specific components on the item detail page) of users, Wen et al. \cite{Micro_Macro2021} propose a Hierarchically Modeling both Micro and Macro behaviors for CVR prediction to address SSB and DS issues by using the abundant supervisory labels from micro and macro behaviors. To models the sequential dependence among multi-step conversions, Adaptive Information Transfer Multi-task (AITM) framework with adaptive information transfer module is developed in \cite{AITM2021} to directly predict the end-to-end conversion probabilities of each step. Besides, causal approaches have also been applied to achieve the debiasing post-click conversion rate estimation lately \cite{Causal_large_2020,Causal_ESCM2_2022,Causal_Estimating_2021,Causal_Enhanced_2021}. However, for the sequential dependence multi-task learning problem, there is rare literature to develop a formalization description.

\section{Conclusions}
In this paper, we propose a sequence dependence multi-task learning framework named as Adaptive Pattern Extraction Multi-task (APEM) framework, which could selectively reconstruct implicit shared representations from a sample-wise view and extract explicit task-specific information in an more efficient way compared with common task-aware tower structure. We accomplish this by involving an Adaptive Sample-wise Representation Generator and a Pattern Selector. For the multi-task learning with dependency generally encountered in E-commence online recommendation, we provide a detail theoretical proof about the dependent relationship from rigorous mathematical perspective. Based on our analysis, we design a dependence task learning loss to complete optimizing object in an unbiased format. The performance gains of APEM compared to several SOTA multi-task approaches on both public and real-world industrial datasets demonstrates its effectiveness and generalization characteristics. Besides, we carefully conduct ablation study, case study, efficiency evaluation and online A/B test to further analyze the contributions from different modules and its applicability for large-scale industrial scenarios.

\section{Appendices}
\subsection{Proof for Theorem \ref{Th-01}}
\begin{proof}
Considering the definitions of the inference and local spaces, and their corresponding expected losses given in (\ref{Eq3-5}) and (\ref{Eq3-6}), we can obtain 
\begin{align}
\label{Eq3-8}
P_{\mathcal{C}}(X,T_{i})=P_\mathcal{D}(X,T_{i}\vert{T_{i-1}=1})
\end{align}
in the sense that the joint distribution of $X$ and $T_{i}$ in $\mathcal{C}$ is equivalent to, under $T_{i-1}=1$, the joint distribution of $X$ and $T_{i}$ in $\mathcal{D}$. 

According to (\ref{Eq3-8}) and the definition of $E_{X,T_{i}\thicksim\mathcal{D}}[L(f_i(X), T_{i})]$, the second term in the right-hand side of (\ref{Eq3-7}) satisfies
\begin{align}
\label{Eq3-9}
E&_{X,T_{i}\thicksim\mathcal{C}}\Big[P_{\mathcal{D}}(T_{i-1}=1)\frac{P_{\mathcal{D}}(T_{i}\vert{X})}{P_{\mathcal{D}}(T_{i},T_{i-1}=1\vert{X})}L(f_i(X),T_{i})\Big]\nonumber\\
=&E_{X,T_{i}\thicksim\mathcal{C}}\Big[\frac{P_{\mathcal{D}}(T_{i-1}=1)}{P_{\mathcal{D}}(T_{i-1}=1\vert{X,T_{i}})}L(f_i(X),T_{i})\Big]\nonumber\\
=&\int_{\mathcal{C}}\frac{P_{\mathcal{D}}(T_{i-1}=1)}{P_{\mathcal{D}}(T_{i-1}=1\vert{x,t_i})}L(f_i(x), t_i)P_\mathcal{C}(x,t_i){\rm d}x{\rm d}t_i\nonumber\\
=&\int_{\mathcal{D}}L(f_i(x), t_i)\frac{P_{\mathcal{D}}(T_{i-1}=1)}{P_{\mathcal{D}}(T_{i-1}=1\vert{x,t_i})}P_\mathcal{D}(x,t_i\vert{T_{i-1}=1}){\rm d}x{\rm d}t_i\nonumber\\
=&\int_{\mathcal{D}}L(f_i(x), t_i)\frac{P_{\mathcal{D}}(T_{i-1}=1)}{P_{\mathcal{D}}(T_{i-1}=1\vert{x,t_i})}\frac{P_\mathcal{D}(x,t_i,T_{i-1}=1)}{P_\mathcal{D}(T_{i-1}=1)}{\rm d}x{\rm d}t_i\nonumber\\
=&\int_{\mathcal{D}}L(f_i(x),t_i)P_\mathcal{D}(x,t_i){\rm d}x{\rm d}t_i\nonumber\\
=&E_{X,T_{i}\thicksim\mathcal{D}}[L(f_i(X),T_{i})].
\end{align}
Therefore, equations (\ref{Eq3-5}), (\ref{Eq3-6}) and (\ref{Eq3-9}) imply that the relationship (\ref{Eq3-7}) holds.
\end{proof}

\subsection{Proof for Theorem \ref{Th-02}}
\begin{proof} 
According to Bayes' theorem, we can obtain the following equality:
\begin{align}
\label{Eq4-n-6}
\frac{P_{\mathcal{D}}(T_{i-1}=1)P_{\mathcal{D}}(T_{i-1}-T_{i}\vert{X})}{P_{\mathcal{D}}(T_{i-1}-T_{i},T_{i-1}=1\vert{X})}=\frac{P_{\mathcal{D}}(T_{i-1}=1)}{P_{\mathcal{D}}(T_{i-1}=1\vert{X,T_{i-1}-T_{i}})}.
\end{align}
Considering the right-hand side of (\ref{Eq4-n-5}) and (\ref{Eq4-n-6}), it leads to
\begin{align}
E&_{X,T_{i-1},T_{i}\thicksim\mathcal{C}}\Bigg[\frac{P_{\mathcal{D}}(T_{i-1}=1)L(f_{i-1}(X)-f_i(X),T_{i-1}-T_{i})}{P_{\mathcal{D}}(T_{i-1}=1\vert{X,T_{i-1}-T_{i}})}\Bigg]\nonumber\\
=&\int_{\mathcal{C}}\Bigg[\frac{P_{\mathcal{D}}(T_{i-1}=1)}{P_{\mathcal{D}}(T_{i-1}=1\vert{x,t_{i-1}-t_{i}})}L(f_{i-1}(x)-f_i(x),t_{i-1}-t_{i})\nonumber\\
&P_{\mathcal{C}}(x,t_{i-1}-t_{i})\Bigg]{\rm d}x{\rm d}t_i{\rm d}t_{i}\nonumber\\
=&\int_{\mathcal{D}}\Bigg[\frac{P_{\mathcal{D}}(T_{i-1}=1)}{P_{\mathcal{D}}(T_{i-1}=1\vert{x,t_{i-1}-t_{i}})}P_{\mathcal{D}}(x,t_{i-1}-t_{i}\vert{T_{i-1}=1})\nonumber\\
&\times{L}(f_{i-1}(x)-f_i(x),t_{i-1}-t_{i})\Bigg]{\rm d}x{\rm d}t_{i-1}{\rm d}t_{i}\nonumber\\
=&\int_{\mathcal{D}}L(f_{i-1}(x)-f_i(x),t_{i-1}-t_{i})P_{\mathcal{D}}(x,t_{i-1}-t_{i}){\rm d}x{\rm d}t_{i-1}{\rm d}t_{i}\nonumber\\
=&\;E_{X,T_{i-1},T_{i}\thicksim\mathcal{D}}[L(f_{i-1}(X)-f_i(X),T_{i-1}-Z)],
\end{align}
which implies that (\ref{Eq4-n-5}) holds. The proof is completed.
\end{proof}

\bibliographystyle{ACM-Reference-Format}
\bibliography{sample-base}


\begin{thebibliography}{32}


\ifx \showCODEN    \undefined \def \showCODEN     #1{\unskip}     \fi
\ifx \showDOI      \undefined \def \showDOI       #1{#1}\fi
\ifx \showISBNx    \undefined \def \showISBNx     #1{\unskip}     \fi
\ifx \showISBNxiii \undefined \def \showISBNxiii  #1{\unskip}     \fi
\ifx \showISSN     \undefined \def \showISSN      #1{\unskip}     \fi
\ifx \showLCCN     \undefined \def \showLCCN      #1{\unskip}     \fi
\ifx \shownote     \undefined \def \shownote      #1{#1}          \fi
\ifx \showarticletitle \undefined \def \showarticletitle #1{#1}   \fi
\ifx \showURL      \undefined \def \showURL       {\relax}        \fi
\providecommand\bibfield[2]{#2}
\providecommand\bibinfo[2]{#2}
\providecommand\natexlab[1]{#1}
\providecommand\showeprint[2][]{arXiv:#2}

\bibitem[Baxter(1997)]%
        {baxter1997bayesian}
\bibfield{author}{\bibinfo{person}{Jonathan Baxter}.}
  \bibinfo{year}{1997}\natexlab{}.
\newblock \showarticletitle{A Bayesian/information theoretic model of learning
  to learn via multiple task sampling}.
\newblock \bibinfo{journal}{\emph{Machine learning}} \bibinfo{volume}{28},
  \bibinfo{number}{1} (\bibinfo{year}{1997}), \bibinfo{pages}{7--39}.
\newblock


\bibitem[Caruana(1997)]%
        {caruana1997multitask}
\bibfield{author}{\bibinfo{person}{Rich Caruana}.}
  \bibinfo{year}{1997}\natexlab{}.
\newblock \showarticletitle{Multitask learning}.
\newblock \bibinfo{journal}{\emph{Machine learning}} \bibinfo{volume}{28},
  \bibinfo{number}{1} (\bibinfo{year}{1997}), \bibinfo{pages}{41--75}.
\newblock


\bibitem[Chen et~al\mbox{.}(2021)]%
        {Chen2021IS}
\bibfield{author}{\bibinfo{person}{Ling Chen}, \bibinfo{person}{Donghui Chen},
  \bibinfo{person}{Fan Yang}, {and} \bibinfo{person}{Jianling Sun}.}
  \bibinfo{year}{2021}\natexlab{}.
\newblock \showarticletitle{Neural episodic control}. In
  \bibinfo{booktitle}{\emph{A deep multi-task representation learning method
  for time series classification and retrieval}}. Information Sciences,
  \bibinfo{pages}{17–32}.
\newblock


\bibitem[Crawshaw(2020)]%
        {crawshaw2020multi}
\bibfield{author}{\bibinfo{person}{Michael Crawshaw}.}
  \bibinfo{year}{2020}\natexlab{}.
\newblock \showarticletitle{Multi-task learning with deep neural networks: A
  survey}.
\newblock \bibinfo{journal}{\emph{arXiv preprint arXiv:2009.09796}}
  (\bibinfo{year}{2020}).
\newblock


\bibitem[Fei et~al\mbox{.}(2021)]%
        {fei2021gemnn}
\bibfield{author}{\bibinfo{person}{Hongliang Fei}, \bibinfo{person}{Jingyuan
  Zhang}, \bibinfo{person}{Xingxuan Zhou}, \bibinfo{person}{Junhao Zhao},
  \bibinfo{person}{Xinyang Qi}, {and} \bibinfo{person}{Ping Li}.}
  \bibinfo{year}{2021}\natexlab{}.
\newblock \showarticletitle{GemNN: gating-enhanced multi-task neural networks
  with feature interaction learning for CTR prediction}. In
  \bibinfo{booktitle}{\emph{Proceedings of the 44th International ACM SIGIR
  Conference on Research and Development in Information Retrieval}}.
  \bibinfo{pages}{2166--2171}.
\newblock


\bibitem[Gu et~al\mbox{.}(2021)]%
        {Causal_Estimating_2021}
\bibfield{author}{\bibinfo{person}{Tiankai Gu}, \bibinfo{person}{Kun Kuang},
  \bibinfo{person}{Hong Zhu}, \bibinfo{person}{Jingjie Li},
  \bibinfo{person}{Zhenhua Dong}, \bibinfo{person}{Wenjie Hu},
  \bibinfo{person}{Zhenguo Li}, \bibinfo{person}{Xiuqiang He}, {and}
  \bibinfo{person}{Yue Liu}.} \bibinfo{year}{2021}\natexlab{}.
\newblock \bibinfo{title}{Estimating true post-click conversion via
  group-stratified counterfactual inference}.
\newblock
\newblock


\bibitem[Guo et~al\mbox{.}(2017)]%
        {guo2017deepfm}
\bibfield{author}{\bibinfo{person}{Huifeng Guo}, \bibinfo{person}{Ruiming
  Tang}, \bibinfo{person}{Yunming Ye}, \bibinfo{person}{Zhenguo Li}, {and}
  \bibinfo{person}{Xiuqiang He}.} \bibinfo{year}{2017}\natexlab{}.
\newblock \showarticletitle{DeepFM: a factorization-machine based neural
  network for CTR prediction}.
\newblock \bibinfo{journal}{\emph{arXiv preprint arXiv:1703.04247}}
  (\bibinfo{year}{2017}).
\newblock


\bibitem[Guo et~al\mbox{.}(2021)]%
        {Causal_Enhanced_2021}
\bibfield{author}{\bibinfo{person}{Siyuan Guo}, \bibinfo{person}{Lixin Zou},
  \bibinfo{person}{Yiding Liu}, \bibinfo{person}{Wenwen Ye},
  \bibinfo{person}{Suqi Cheng}, \bibinfo{person}{Shuaiqiang Wang},
  \bibinfo{person}{Hechang Chen}, \bibinfo{person}{Dawei Yin}, {and}
  \bibinfo{person}{Yi Chang}.} \bibinfo{year}{2021}\natexlab{}.
\newblock \showarticletitle{Enhanced doubly robust learning for debiasing
  post-click conversion rate estimation}. In
  \bibinfo{booktitle}{\emph{Proceedings of the 44th International ACM SIGIR
  Conference on Research and Development in Information Retrieval}}.
  \bibinfo{pages}{275--284}.
\newblock


\bibitem[Hazimeh et~al\mbox{.}(2021)]%
        {2021DSelect}
\bibfield{author}{\bibinfo{person}{H. Hazimeh}, \bibinfo{person}{Z. Zhao},
  \bibinfo{person}{A. Chowdhery}, \bibinfo{person}{M. Sathiamoorthy}, {and}
  \bibinfo{person}{E.~H. Chi}.} \bibinfo{year}{2021}\natexlab{}.
\newblock \showarticletitle{DSelect-k: Differentiable Selection in the Mixture
  of Experts with Applications to Multi-Task Learning}.
\newblock  (\bibinfo{year}{2021}).
\newblock


\bibitem[Jacobs et~al\mbox{.}(1991)]%
        {jacobs1991adaptive}
\bibfield{author}{\bibinfo{person}{Robert~A Jacobs}, \bibinfo{person}{Michael~I
  Jordan}, \bibinfo{person}{Steven~J Nowlan}, {and} \bibinfo{person}{Geoffrey~E
  Hinton}.} \bibinfo{year}{1991}\natexlab{}.
\newblock \showarticletitle{Adaptive mixtures of local experts}.
\newblock \bibinfo{journal}{\emph{Neural computation}} \bibinfo{volume}{3},
  \bibinfo{number}{1} (\bibinfo{year}{1991}), \bibinfo{pages}{79--87}.
\newblock


\bibitem[Kingma and Ba(2014)]%
        {kingma2014adam}
\bibfield{author}{\bibinfo{person}{Diederik~P Kingma} {and}
  \bibinfo{person}{Jimmy Ba}.} \bibinfo{year}{2014}\natexlab{}.
\newblock \showarticletitle{Adam: A method for stochastic optimization}.
\newblock \bibinfo{journal}{\emph{arXiv preprint arXiv:1412.6980}}
  (\bibinfo{year}{2014}).
\newblock


\bibitem[Lee et~al\mbox{.}(2019)]%
        {lee2019set}
\bibfield{author}{\bibinfo{person}{Juho Lee}, \bibinfo{person}{Yoonho Lee},
  \bibinfo{person}{Jungtaek Kim}, \bibinfo{person}{Adam Kosiorek},
  \bibinfo{person}{Seungjin Choi}, {and} \bibinfo{person}{Yee~Whye Teh}.}
  \bibinfo{year}{2019}\natexlab{}.
\newblock \showarticletitle{Set transformer: A framework for attention-based
  permutation-invariant neural networks}. In
  \bibinfo{booktitle}{\emph{International Conference on Machine Learning}}.
  PMLR, \bibinfo{pages}{3744--3753}.
\newblock


\bibitem[Lerman(1980)]%
        {lerman1980fitting}
\bibfield{author}{\bibinfo{person}{PM Lerman}.}
  \bibinfo{year}{1980}\natexlab{}.
\newblock \showarticletitle{Fitting segmented regression models by grid
  search}.
\newblock \bibinfo{journal}{\emph{Journal of the Royal Statistical Society:
  Series C (Applied Statistics)}} \bibinfo{volume}{29}, \bibinfo{number}{1}
  (\bibinfo{year}{1980}), \bibinfo{pages}{77--84}.
\newblock


\bibitem[Ma et~al\mbox{.}(2018b)]%
        {ma2018modeling}
\bibfield{author}{\bibinfo{person}{Jiaqi Ma}, \bibinfo{person}{Zhe Zhao},
  \bibinfo{person}{Xinyang Yi}, \bibinfo{person}{Jilin Chen},
  \bibinfo{person}{Lichan Hong}, {and} \bibinfo{person}{Ed~H Chi}.}
  \bibinfo{year}{2018}\natexlab{b}.
\newblock \showarticletitle{Modeling task relationships in multi-task learning
  with multi-gate mixture-of-experts}. In \bibinfo{booktitle}{\emph{Proceedings
  of the 24th ACM SIGKDD International Conference on Knowledge Discovery \&
  Data Mining}}. \bibinfo{pages}{1930--1939}.
\newblock


\bibitem[Ma et~al\mbox{.}(2018a)]%
        {ESMM2018}
\bibfield{author}{\bibinfo{person}{Xiao Ma}, \bibinfo{person}{Liqin Zhao},
  \bibinfo{person}{Guan Huang}, \bibinfo{person}{Zhi Wang},
  \bibinfo{person}{Zelin Hu}, \bibinfo{person}{Xiaoqiang Zhu}, {and}
  \bibinfo{person}{Kun Gai}.} \bibinfo{year}{2018}\natexlab{a}.
\newblock \showarticletitle{Entire space multi-task model: An effective
  approach for estimating post-click conversion rate}. In
  \bibinfo{booktitle}{\emph{The 41st International ACM SIGIR Conference on
  Research \& Development in Information Retrieval}}.
  \bibinfo{pages}{1137--1140}.
\newblock


\bibitem[Misra et~al\mbox{.}(2016)]%
        {misra2016cross}
\bibfield{author}{\bibinfo{person}{Ishan Misra}, \bibinfo{person}{Abhinav
  Shrivastava}, \bibinfo{person}{Abhinav Gupta}, {and} \bibinfo{person}{Martial
  Hebert}.} \bibinfo{year}{2016}\natexlab{}.
\newblock \showarticletitle{Cross-stitch networks for multi-task learning}. In
  \bibinfo{booktitle}{\emph{Proceedings of the IEEE conference on computer
  vision and pattern recognition}}. \bibinfo{pages}{3994--4003}.
\newblock


\bibitem[O'Brien et~al\mbox{.}(2021)]%
        {o2021analysis}
\bibfield{author}{\bibinfo{person}{Conor O'Brien}, \bibinfo{person}{Kin~Sum
  Liu}, \bibinfo{person}{James Neufeld}, \bibinfo{person}{Rafael Barreto},
  {and} \bibinfo{person}{Jonathan~J Hunt}.} \bibinfo{year}{2021}\natexlab{}.
\newblock \showarticletitle{An Analysis Of Entire Space Multi-Task Models For
  Post-Click Conversion Prediction}. In \bibinfo{booktitle}{\emph{Fifteenth ACM
  Conference on Recommender Systems}}. \bibinfo{pages}{613--619}.
\newblock


\bibitem[Pritzel et~al\mbox{.}(2017)]%
        {pritzel2017neural}
\bibfield{author}{\bibinfo{person}{Alexander Pritzel}, \bibinfo{person}{Benigno
  Uria}, \bibinfo{person}{Sriram Srinivasan},
  \bibinfo{person}{Adria~Puigdomenech Badia}, \bibinfo{person}{Oriol Vinyals},
  \bibinfo{person}{Demis Hassabis}, \bibinfo{person}{Daan Wierstra}, {and}
  \bibinfo{person}{Charles Blundell}.} \bibinfo{year}{2017}\natexlab{}.
\newblock \showarticletitle{Neural episodic control}. In
  \bibinfo{booktitle}{\emph{International Conference on Machine Learning}}.
  PMLR, \bibinfo{pages}{2827--2836}.
\newblock


\bibitem[Ruder et~al\mbox{.}(2019)]%
        {ruder2019latent}
\bibfield{author}{\bibinfo{person}{Sebastian Ruder}, \bibinfo{person}{Joachim
  Bingel}, \bibinfo{person}{Isabelle Augenstein}, {and} \bibinfo{person}{Anders
  S{\o}gaard}.} \bibinfo{year}{2019}\natexlab{}.
\newblock \showarticletitle{Latent multi-task architecture learning}. In
  \bibinfo{booktitle}{\emph{Proceedings of the AAAI Conference on Artificial
  Intelligence}}, Vol.~\bibinfo{volume}{33}. \bibinfo{pages}{4822--4829}.
\newblock


\bibitem[Shen et~al\mbox{.}(2021)]%
        {shen2021variational}
\bibfield{author}{\bibinfo{person}{Jiayi Shen}, \bibinfo{person}{Xiantong
  Zhen}, \bibinfo{person}{Marcel Worring}, {and} \bibinfo{person}{Ling Shao}.}
  \bibinfo{year}{2021}\natexlab{}.
\newblock \showarticletitle{Variational multi-task learning with gumbel-softmax
  priors}.
\newblock \bibinfo{journal}{\emph{Advances in Neural Information Processing
  Systems}}  \bibinfo{volume}{34} (\bibinfo{year}{2021}),
  \bibinfo{pages}{21031--21042}.
\newblock


\bibitem[Stickland and Murray(2019)]%
        {stickland2019bert}
\bibfield{author}{\bibinfo{person}{Asa~Cooper Stickland} {and}
  \bibinfo{person}{Iain Murray}.} \bibinfo{year}{2019}\natexlab{}.
\newblock \showarticletitle{Bert and pals: Projected attention layers for
  efficient adaptation in multi-task learning}. In
  \bibinfo{booktitle}{\emph{International Conference on Machine Learning}}.
  PMLR, \bibinfo{pages}{5986--5995}.
\newblock


\bibitem[Tang et~al\mbox{.}(2020)]%
        {tang2020progressive}
\bibfield{author}{\bibinfo{person}{Hongyan Tang}, \bibinfo{person}{Junning
  Liu}, \bibinfo{person}{Ming Zhao}, {and} \bibinfo{person}{Xudong Gong}.}
  \bibinfo{year}{2020}\natexlab{}.
\newblock \showarticletitle{Progressive layered extraction (PLE): A novel
  multi-task learning (MTL) model for personalized recommendations}. In
  \bibinfo{booktitle}{\emph{Fourteenth ACM Conference on Recommender Systems}}.
  \bibinfo{pages}{269--278}.
\newblock


\bibitem[Vafaeikia et~al\mbox{.}(2020)]%
        {vafaeikia2020brief}
\bibfield{author}{\bibinfo{person}{Partoo Vafaeikia},
  \bibinfo{person}{Khashayar Namdar}, {and} \bibinfo{person}{Farzad Khalvati}.}
  \bibinfo{year}{2020}\natexlab{}.
\newblock \showarticletitle{A Brief Review of Deep Multi-task Learning and
  Auxiliary Task Learning}.
\newblock \bibinfo{journal}{\emph{arXiv preprint arXiv:2007.01126}}
  (\bibinfo{year}{2020}).
\newblock


\bibitem[Vandenhende et~al\mbox{.}(2020)]%
        {vandenhende2020revisiting}
\bibfield{author}{\bibinfo{person}{Simon Vandenhende},
  \bibinfo{person}{Stamatios Georgoulis}, \bibinfo{person}{Marc Proesmans},
  \bibinfo{person}{Dengxin Dai}, {and} \bibinfo{person}{Luc Van~Gool}.}
  \bibinfo{year}{2020}\natexlab{}.
\newblock \showarticletitle{Revisiting multi-task learning in the deep learning
  era}.
\newblock \bibinfo{journal}{\emph{arXiv preprint arXiv:2004.13379}}
  \bibinfo{volume}{2} (\bibinfo{year}{2020}).
\newblock


\bibitem[Wang et~al\mbox{.}(2022b)]%
        {wang2022enhancing}
\bibfield{author}{\bibinfo{person}{Fangye Wang}, \bibinfo{person}{Yingxu Wang},
  \bibinfo{person}{Dongsheng Li}, \bibinfo{person}{Hansu Gu},
  \bibinfo{person}{Tun Lu}, \bibinfo{person}{Peng Zhang}, {and}
  \bibinfo{person}{Ning Gu}.} \bibinfo{year}{2022}\natexlab{b}.
\newblock \showarticletitle{Enhancing CTR Prediction with Context-Aware Feature
  Representation Learning}.
\newblock \bibinfo{journal}{\emph{arXiv preprint arXiv:2204.08758}}
  (\bibinfo{year}{2022}).
\newblock


\bibitem[Wang et~al\mbox{.}(2022a)]%
        {Causal_ESCM2_2022}
\bibfield{author}{\bibinfo{person}{Hao Wang}, \bibinfo{person}{Tai-Wei Chang},
  \bibinfo{person}{Tianqiao Liu}, \bibinfo{person}{Jianmin Huang},
  \bibinfo{person}{Zhichao Chen}, \bibinfo{person}{Chao Yu},
  \bibinfo{person}{Ruopeng Li}, {and} \bibinfo{person}{Wei Chu}.}
  \bibinfo{year}{2022}\natexlab{a}.
\newblock \showarticletitle{$\text{ESCM}^2$: Entire Space Counterfactual
  Multi-Task Model for Post-Click Conversion Rate Estimation}.
\newblock \bibinfo{journal}{\emph{arXiv preprint arXiv:2204.05125}}
  (\bibinfo{year}{2022}).
\newblock


\bibitem[Wang et~al\mbox{.}(2020)]%
        {wang2020k}
\bibfield{author}{\bibinfo{person}{Ruize Wang}, \bibinfo{person}{Duyu Tang},
  \bibinfo{person}{Nan Duan}, \bibinfo{person}{Zhongyu Wei},
  \bibinfo{person}{Xuanjing Huang}, \bibinfo{person}{Guihong Cao},
  \bibinfo{person}{Daxin Jiang}, \bibinfo{person}{Ming Zhou}, {et~al\mbox{.}}}
  \bibinfo{year}{2020}\natexlab{}.
\newblock \showarticletitle{K-adapter: Infusing knowledge into pre-trained
  models with adapters}.
\newblock \bibinfo{journal}{\emph{arXiv preprint arXiv:2002.01808}}
  (\bibinfo{year}{2020}).
\newblock


\bibitem[Wen et~al\mbox{.}(2021)]%
        {Micro_Macro2021}
\bibfield{author}{\bibinfo{person}{Hong Wen}, \bibinfo{person}{Jing Zhang},
  \bibinfo{person}{Fuyu Lv}, \bibinfo{person}{Wentian Bao},
  \bibinfo{person}{Tianyi Wang}, {and} \bibinfo{person}{Zulong Chen}.}
  \bibinfo{year}{2021}\natexlab{}.
\newblock \showarticletitle{Hierarchically modeling micro and macro behaviors
  via multi-task learning for conversion rate prediction}. In
  \bibinfo{booktitle}{\emph{Proceedings of the 44th International ACM SIGIR
  Conference on Research and Development in Information Retrieval}}.
  \bibinfo{pages}{2187--2191}.
\newblock


\bibitem[Wen et~al\mbox{.}(2020)]%
        {Entire_SIGIR_2020}
\bibfield{author}{\bibinfo{person}{Hong Wen}, \bibinfo{person}{Jing Zhang},
  \bibinfo{person}{Yuan Wang}, \bibinfo{person}{Fuyu Lv},
  \bibinfo{person}{Wentian Bao}, \bibinfo{person}{Quan Lin}, {and}
  \bibinfo{person}{Keping Yang}.} \bibinfo{year}{2020}\natexlab{}.
\newblock \showarticletitle{Entire space multi-task modeling via post-click
  behavior decomposition for conversion rate prediction}. In
  \bibinfo{booktitle}{\emph{Proceedings of the 43rd International ACM SIGIR
  conference on research and development in Information Retrieval}}.
  \bibinfo{pages}{2377--2386}.
\newblock


\bibitem[Xi et~al\mbox{.}(2021)]%
        {AITM2021}
\bibfield{author}{\bibinfo{person}{Dongbo Xi}, \bibinfo{person}{Zhen Chen},
  \bibinfo{person}{Peng Yan}, \bibinfo{person}{Yinger Zhang},
  \bibinfo{person}{Yongchun Zhu}, \bibinfo{person}{Fuzhen Zhuang}, {and}
  \bibinfo{person}{Yu Chen}.} \bibinfo{year}{2021}\natexlab{}.
\newblock \showarticletitle{Modeling the sequential dependence among audience
  multi-step conversions with multi-task learning in targeted display
  advertising}. In \bibinfo{booktitle}{\emph{Proceedings of the 27th ACM SIGKDD
  Conference on Knowledge Discovery \& Data Mining}}.
  \bibinfo{pages}{3745--3755}.
\newblock


\bibitem[Zhang et~al\mbox{.}(2020)]%
        {Causal_large_2020}
\bibfield{author}{\bibinfo{person}{Wenhao Zhang}, \bibinfo{person}{Wentian
  Bao}, \bibinfo{person}{Xiao-Yang Liu}, \bibinfo{person}{Keping Yang},
  \bibinfo{person}{Quan Lin}, \bibinfo{person}{Hong Wen}, {and}
  \bibinfo{person}{Ramin Ramezani}.} \bibinfo{year}{2020}\natexlab{}.
\newblock \showarticletitle{Large-scale causal approaches to debiasing
  post-click conversion rate estimation with multi-task learning}. In
  \bibinfo{booktitle}{\emph{Proceedings of The Web Conference 2020}}.
  \bibinfo{pages}{2775--2781}.
\newblock


\bibitem[Zhao et~al\mbox{.}(2019)]%
        {zhao2019multiple}
\bibfield{author}{\bibinfo{person}{Jiejie Zhao}, \bibinfo{person}{Bowen Du},
  \bibinfo{person}{Leilei Sun}, \bibinfo{person}{Fuzhen Zhuang},
  \bibinfo{person}{Weifeng Lv}, {and} \bibinfo{person}{Hui Xiong}.}
  \bibinfo{year}{2019}\natexlab{}.
\newblock \showarticletitle{Multiple relational attention network for
  multi-task learning}. In \bibinfo{booktitle}{\emph{Proceedings of the 25th
  ACM SIGKDD International Conference on Knowledge Discovery \& Data Mining}}.
  \bibinfo{pages}{1123--1131}.
\newblock


\end{thebibliography}










\end{document}